\documentclass[]{fairmeta}

\usepackage{makecell}
\usepackage[dvipsnames]{xcolor}
\usepackage{arydshln}
\usepackage{tikz}
\usepackage{adjustbox}
\usepackage{wrapfig}
\usepackage{algorithm}
\usepackage{algorithmic}
\usepackage{bbm}
\usepackage{epigraph}
\usepackage{threeparttable}
\usepackage{orcidlink}
\usepackage{amsmath,amssymb}
\usepackage[sc]{mathpazo}
\usepackage{float}
\usepackage{todonotes}

\usepackage[utf8]{inputenc} 
\usepackage[T1]{fontenc}    
\usepackage{hyperref}       
\usepackage{url}            
\usepackage{booktabs}       
\usepackage{amsfonts}       
\usepackage{nicefrac}       
\usepackage{microtype}      
\usepackage{xcolor}         

\usepackage{orcidlink}
\usepackage{tikz}
\usepackage{adjustbox}
\usepackage{wrapfig}
\usepackage{makecell}
\usepackage{multirow}
\usepackage{algorithm}
\usepackage{algorithmic}
\usepackage{wrapfig}
\usepackage{bbm}
\usepackage{tikz}
\usepackage{epigraph} 
\usepackage{threeparttable}
\usepackage{subcaption}
\usepackage{soul}
\usepackage{wrapfig}
\usepackage{amssymb}
\usepackage{amsmath}
\usepackage{xspace}
\newcommand{\ie}{i.e.,\xspace}
\newcommand{\eg}{e.g.,\xspace}
\newcommand{\etc}{etc.\xspace}
\usepackage{arydshln}
\usetikzlibrary{arrows.meta}
\usepackage{tcolorbox}
\usepackage{listings}
\usepackage{float}
\usepackage[dvipsnames]{xcolor}
\usepackage[numbers]{natbib}
\let\citep\cite

\usetikzlibrary{arrows.meta}

\definecolor{lightgreen}{HTML}{E6FFE6}

\RequirePackage{xspace}
\makeatletter
\DeclareRobustCommand\onedot{\futurelet\@let@token\@onedot}
\def\@onedot{\ifx\@let@token.\else.\null\fi\xspace}

\def\eg{\emph{e.g}\onedot}

\def\ie{\emph{i.e}\onedot}

\def\etc{\emph{etc}\onedot}

\makeatother

\crefname{figure}{Fig.}{Fig.}

\crefname{table}{Tab.}{Tab.}

\crefname{section}{Sec.}{Sec.}

\crefname{appendix}{Appendix}{Appendices}

\title{YoCausal: How Far is Video Generation from World Model? A Causality Perspective}

\author[1,2,*]{You-Zhe Xie}
\author[1,*]{Yu-Hsuan Li}
\author[1]{Jie-Ying Lee}
\author[2]{Kaipeng Zhang}
\author[1,\dagger]{Yu-Lun Liu}
\author[2,\dagger]{Zhixiang Wang}

\affiliation[1]{National Yang Ming Chiao Tung University}
\affiliation[2]{Alaya Lab}

\contribution[*]{Equal contribution}
\contribution[\dagger]{Corresponding authors}

\abstract{
As video diffusion models (VDMs) advance toward world models, a key question arises: do they truly understand causality, or merely overfit to statistical temporal patterns? Existing benchmarks mostly rely on synthetic data, limiting real-world generalization due to the sim-to-real gap. We present YoCausal, a two-level benchmark inspired by the Violation of Expectation (VoE) paradigm from cognitive science. By temporally reversing real-world videos at zero cost as natural counterfactual samples, YoCausal establishes an arbitrarily extensible evaluation protocol. Level 1 introduces the Reverse Surprise Index (RSI), quantifying arrow-of-time perception via denoising loss. Level 2 introduces the Causality Cognition Index (CCI), which leverages a VLM to stratify datasets into causal and non-causal subsets, disentangling genuine causal reasoning from temporal bias. Evaluation of 13 state-of-the-art VDMs reveals that perceiving the arrow of time does not imply understanding causality, and a significant gap persists relative to human-level causal cognition.

\github{\url{https://www.youzhexie.me/papers/YoCausal/index.html}}
\correspondence{\email{yulunliu@cs.nycu.edu.tw}, \email{wangzx1994@gmail.com}}
}
\date{\today}

\begin{document}
\maketitle

\begin{figure}[!h]
    \centering
  \begin{tikzpicture}
      \node[anchor=south west, inner sep=0] (image) at (0,0) {
          \includegraphics[width=0.96\linewidth]{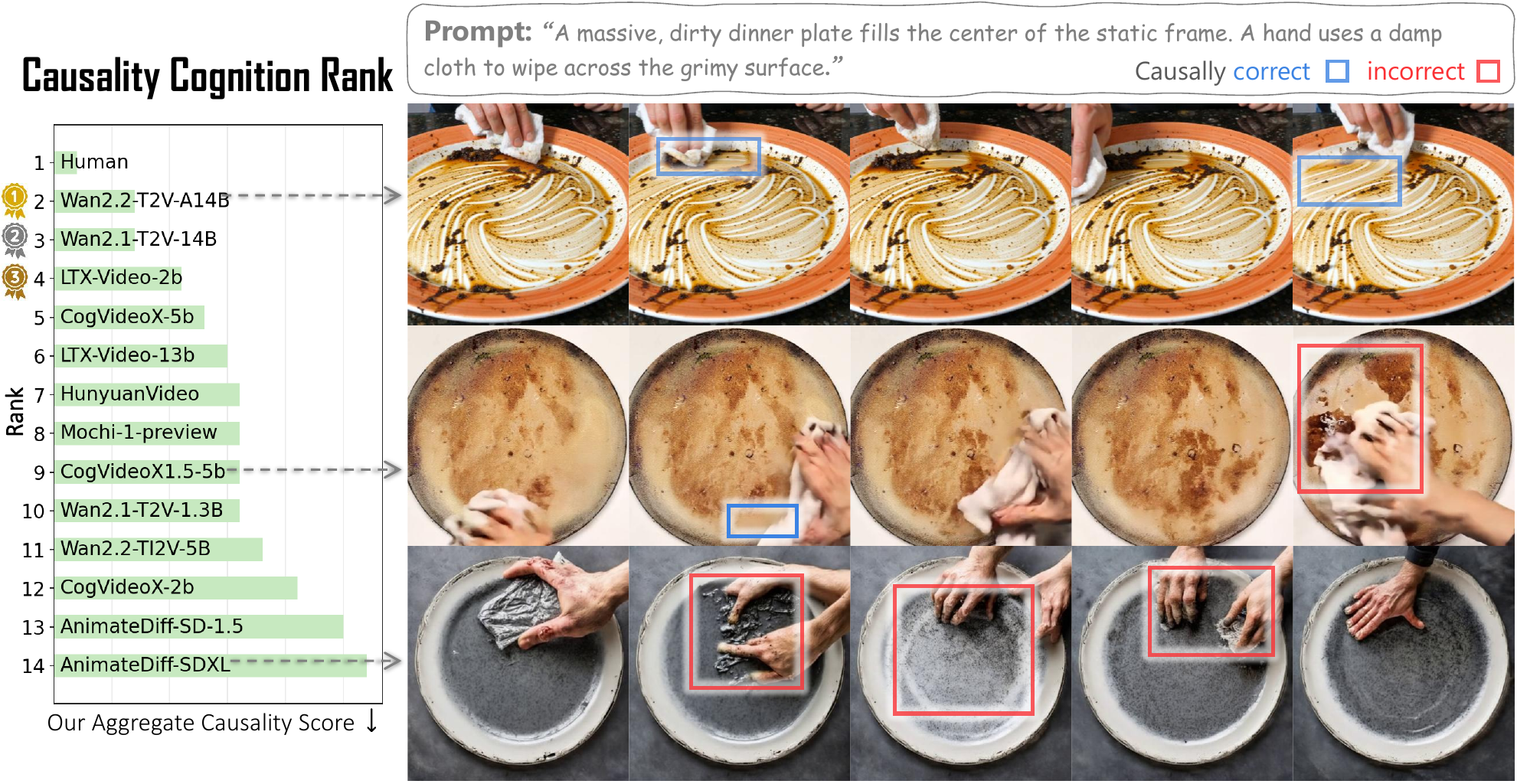}
      };
      \fill[white] (3.83, 0.0) rectangle (4.3, 0.9);
      \node at (3.95, 0.65) {\tiny$\downarrow$};
  \end{tikzpicture}
  \caption{
      \textbf{Validation of the YoCausal benchmark.}
      \textbf{(Left)} We evaluate and rank 13 state-of-the-art Video Diffusion Models alongside a human baseline based on our Causality Score (lower is better).
      \textbf{(Right)} Visualizations of a causal event (``\emph{wiping a dirty plate}'') strongly align with our quantitative rankings.
      The top-ranked VDM, Wan2.2-A14B~\cite{wan2025wan}, successfully captures the cause-and-effect relationship by progressively removing the dirt (\textcolor{Blue}{\textbf{blue boxes}}).
      In contrast, mid- to low-ranked models like CogVideoX1.5-5B~\cite{hong2022cogvideo,yang2024cogvideox} and AnimateDiff-SDXL~\cite{guo2023animatediff} struggle to understand the underlying causal logic.
      They show severe causal errors, such as dirt reappearing or smearing in an illogical way (\textcolor{red}{\textbf{red boxes}}).
      This shows our benchmark probes causal cognition of video generation model.
  }
  \label{fig:teaser}
\end{figure}

\section{Introduction}
\label{sec:intro}

\setlength{\epigraphwidth}{0.75\textwidth}
\epigraph{\scriptsize{``Welcome to the exploration of causal video generation.''}}
{\scriptsize{---\textbf{YoCausal}, {Derived from \textit{Yōkoso}  (``welcome'' in Japanese).}}}


A long-standing aspiration of AI is to build machines that truly model the world~\citep{lecun2022path,ha2018world,matsuo2022deep,hu2023gaia,bruce2024genie,lin2023learning}. One important ability of such \emph{world models} is to capture \emph{causality}\footnote{In this paper, we focus on intuitively observable cause-and-effect (\ie event $A$ leads to event $B$), rather than structural causal models or interventions~\citep{neuberg2003causality,didelez2001causality}.}.

\emph{Video generation models} have been regarded as a direct path to world modeling~\citep{liu2024sora,esser2023structure,qin2024worldsimbench,li2025worldmodelbench,zhu2024sora}. Trained on vast real-world data, they can produce highly realistic video. However, a fundamental question remains: \textbf{do current video generation models actually understand causality?}


Previous research on ``world knowledge'' has focused on adherence to physical laws~\citep{riochet2018intphys,bordes2025intphys,yuan2025likephys,motamed2026generative,wu2016physics,bear2021physion,kang2024far,bansal2024videophy,bansal2025videophy,meng2024towards}. However, a genuine world model must go beyond physics to comprehend broader causality. Existing physics benchmarks face a limitation: to isolate physical variables, they rely on synthetic data or small collections of controlled laboratory recordings, creating a \emph{sim-to-real gap} that limits assessment of real-world generalization.

\begin{figure*}[t]
    \centering
    \includegraphics[width=1.0\linewidth]{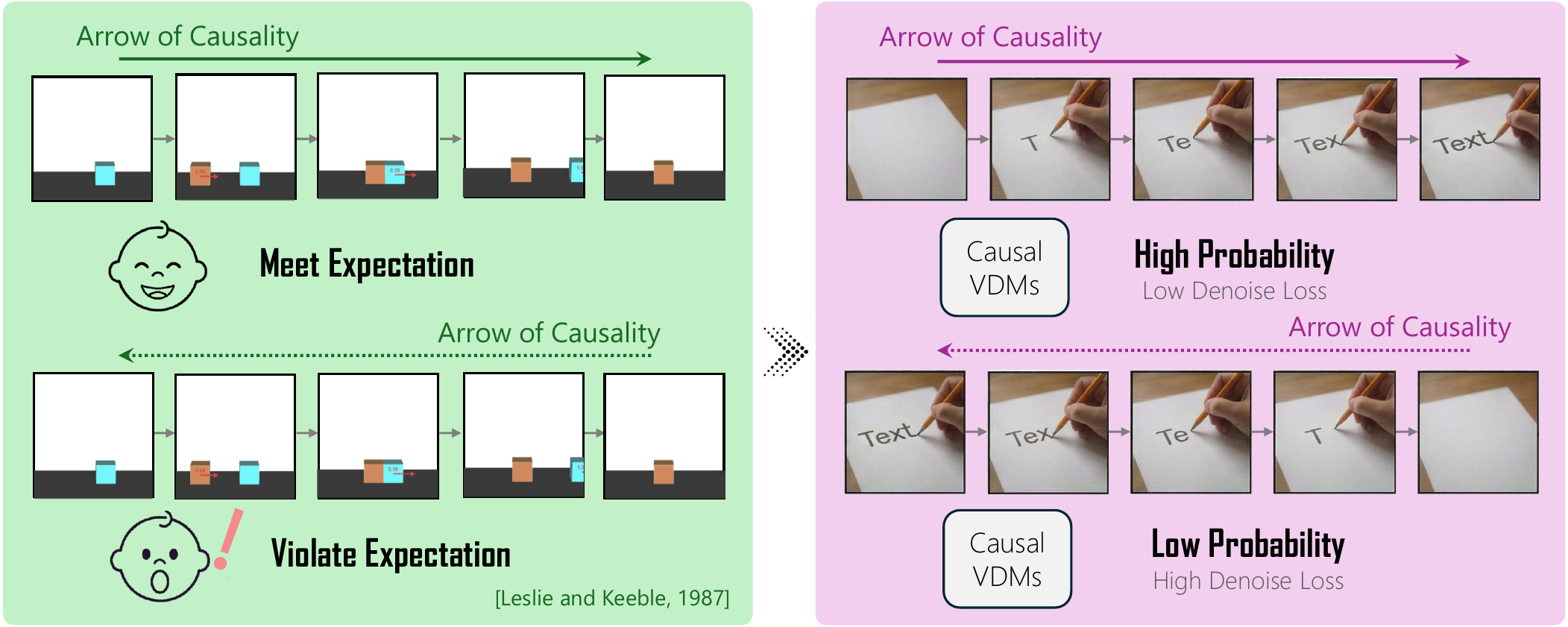}
    \caption{
    \textbf{Conceptual overview of YoCausal benchmark.}
    We draw inspiration from the \textbf{Violation of Expectation (VoE)} paradigm in cognitive science.
    \textbf{(Left)} Infants show surprise when seeing videos played in reverse (bottom), violating their intuitive causal cognition~\cite{leslie1987six} ($\dashrightarrow$).
    \textbf{(Right)} We transfer this paradigm to generative models: treating the learned data distribution as the model's cognition, a causally-aware VDM should assign lower probability (higher denoising loss) to counterfactual reversed videos than to forward ones. This elegant analogy allows us to benchmark causal understanding using arbitrarily scalable real-world videos.
    }
    \label{fig:motivation_fig}
\end{figure*}

To bridge this gap, we draw on the \emph{Violation of Expectation (VoE)} paradigm from cognitive science~\citep{margoni2024violation}. In a seminal study (\cref{fig:motivation_fig}), Leslie and Keeble~\citep{leslie1987six} assessed whether infants perceive causality by showing temporally reversed videos

Building on this, we propose \textbf{YoCausal}, a two-level benchmark for
causal cognition in VDMs. At Level~1, the \emph{Reverse Surprise Index (RSI)}
measures the proportion a model assigns lower likelihood to the reversed than the
forward video. However, RSI alone cannot isolate causal cognition, as RSI may come from both the \emph{arrow of time} and causality. Level~2 therefore introduces the
\emph{Causality Cognition Index (CCI)}, which contrasts RSI on causal versus
non-causal videos: a causally aware model should be more surprised by reversed
causal events.

A key advantage is our \emph{arbitrarily extensible dataset} design. Any real-world video can be temporally reversed at zero cost to produce a counterfactual sample, freeing the benchmark from the fixed synthetic scenes or controlled recordings of prior work. This bridges the sim-to-real gap. We further provide performance of human annotators as a reference for model performance.

Comprehensive evaluation across 13 state-of-the-art VDMs reveals four key insights: (1)~while some advanced models exhibit preliminary causality cognition, a significant human-model gap remains; (2)~perceiving the arrow of time is not equivalent to understanding causality; (3)~causal cognition correlates partially with intuitive physics but almost not with aesthetic quality and (4)~scaling parameters and advancing architectures improve causal cognition, indicating that scaling laws\citep{yin2025towards,kaplan2020scaling} extend to this higher-order reasoning.

In summary, our contributions are as follows:
\begin{itemize}
    \item The first likelihood-based, VoE-grounded causality benchmark for VDMs, built on a
    scalable real-world dataset free from sim-to-real gaps.
    \item A cognitive-science-grounded two-level framework that
    disentangles arrow-of-time perception from causal cognition.
    \item Evidence that current open-source VDMs lack
    causal understanding, revealing a critical gap toward
    world models and providing guidance.
\end{itemize}
\vspace{-1.5em}
\section{Related Work}
\label{sec:related}

\paragraph{Video Diffusion Models.}
Video synthesis has scaled from UNet~\citep{ho2022video, wang2023modelscope,
blattmann2023stable, singer2022make}
to DiT~\citep{peebles2023scalable} backbones~\citep{zheng2024open,
peng2025open} and commercial systems~\citep{brooks2024video, polyak2024movie}
with debated zero-shot reasoning~\citep{wiedemer2025video, guo2026video},
motivating VDMs as \emph{world models}~\citep{lecun2022path, ha2018world} for
interactive simulation and planning~\citep{bruce2024genie, yang2023learning,
valevski2024diffusion, agarwal2025cosmos, hafner2020mastering,
hafner2019learning, du2023video}. A world model differs by its causal structure~\citep{pearl2003causality,
scholkopf2021toward}, which robust generalization under distribution shift requires~\citep{richens2024robust}. Whether VDMs acquire it remains
open~\citep{kang2024far, motamed2026generative, bai2025impossible,
zhang2026world}, and memorization of content and
motion~\citep{carlini2023extracting, somepalli2023diffusion, rahman2025frame,
chen2024investigating} makes realistic rendering weak evidence of
understanding.

\paragraph{Video Generation Evaluation.}
Evaluation has evolved from distribution metrics~\citep{unterthiner2018towards,
ge2024content} to multi-dimensional suites~\citep{huang2024vbench,
zheng2025vbench, liu2024evalcrafter, sun2025t2v} and
temporal, physical, and world-model benchmarks~\citep{liu2024tempcompass,
cai2024temporalbench, yuan2024chronomagic, bansal2024videophy,
meng2024towards, guo2025t2vphysbench, chen2025physical,
upadhyay2026worldbench, thozhiyoor2026principia, han2026oscbench,
duan2025worldscore}; concurrent work frames evaluation causally via
prompt, appearance, viewpoint, or action interventions (What-If
World~\citep{cai2026if}, CRONOS~\citep{begiristain2026cronos},
Omni-WorldBench~\citep{wu2026omni}). These score \emph{generated} videos with
VLM judges~\citep{he2024videoscore} that inherit known
biases~\citep{zheng2023judging}, or with pixel
comparisons~\citep{zhang2025morpheus}.

\paragraph{Intuitive Physics and VoE.}
Rooted in perceptual causality~\citep{michotte2017perception} and infant core
knowledge, including causal launching~\citep{baillargeon1985object,
spelke1992origins, stahl2015observing, spelke2007core, teglas2011pure,
baillargeon2004infants, ullman2017mind, oakes1990infant, saxe2006perception},
the \emph{violation-of-expectation} (VoE) paradigm~\citep{margoni2024violation}
measures cognition via surprise at counterfactuals, also in adults viewing
naturalistic video~\citep{plikat2025hierarchical}; we define surprise as an
observation's improbability under an internal model~\citep{friston2010free}
rather than the belief update it induces~\citep{itti2009bayesian}. Since
human-like intelligence requires intuitive causal
theories~\citep{lake2017building}, VoE has been applied to discriminative
models on synthetic benchmarks~\citep{riochet2021intphys, bordes2025intphys,
dasgupta2021avoe, smith2019modeling, baradel2019cophy, gandhi2021baby,
shu2021agent, garrido2025intuitive, piloto2022intuitive,
battaglia2013simulation}, to latent world-model prediction
error~\citep{maes2026leworldmodel}, and to generative
likelihoods~\citep{buschoff2023acquisition}. LikePhys~\citep{yuan2025likephys}
extends it to VDMs via the denoising loss as an ELBO-grounded likelihood
proxy~\citep{chao2022denoising, li2023your, clark2023text, song2020score,
song2021maximum, kingma2021variational, nichol2021improved,
kingma2023understanding}, best used in relative form since such likelihoods
can be dominated by low-level statistics~\citep{nalisnick2018deep,
ren2019likelihood, graham2023denoising}; related probes use a generator's
flow residual~\citep{hassan2026proprio} or internal
states~\citep{esmati2026invisible}.

\paragraph{Arrow of Time and Causality in Video Understanding.}
Temporal directionality~\citep{layzer1975arrow} serves as a self-supervised
signal~\citep{misra2016shuffle, lee2017unsupervised, wei2018learning,
xu2019self, pickup2014seeing} yet remains hard for multimodal
models~\citep{bagad2023test, shangguan2025tomato, zhang2024vinoground,
xue2025seeing, cores2024tvbench, li2024vitatecs, han2026tracing,
yu2026arrowgev}, with VLMs near chance on forward/backward
stimuli from human psychophysics~\citep{matta2026way, hanyu2023ready}. Video
understanding~\citep{fu2025video, li2024mvbench, patraucean2023perception,
chandrasegaran2024hourvideo, li2024eyes} and causal
reasoning~\citep{liu2025can} benchmarks span synthetic~\citep{yi2019clevrer,
ates2022craft} and real-world~\citep{li2022representation, xiao2021next,
foss2025causalvqa, chen2025countervqa, chi2024unveiling,
wang2025timecausality} video QA and language
counterfactuals~\citep{zevcevic2023causal, kiciman2023causal,
jin2023cladder}, with Bayesian surprise recently brought to video
LLMs~\citep{ravi2026spike}. All target \emph{discriminative} tasks; on the
generative side, prior work steers VDMs backward in
time~\citep{feng2024explorative, liu2026can} or diagnoses under-developed
irreversible processes in generated videos~\citep{xu2026diagnosing}.

\paragraph{Positioning.}
Existing evaluations, including the concurrent What-If World, CRONOS, and
Omni-WorldBench, either judge generated videos, entangling causality with
appearance, or target discriminative models and synthetic physics. YoCausal
fills this gap by measuring whether a \emph{generative} VDM's learned prior
has internalized causal structure, read directly from its own likelihood on
real-world videos without generation or an external judge. To our knowledge, it is the
first likelihood-based, VoE-grounded causality benchmark for VDMs on
real-world videos.

\section{Method}
\label{sec:method}

\begin{table*}[t]
\centering
\caption{
\textbf{Comparison with existing physics-law evaluation benchmarks.}
Prior benchmarks rely on synthetic data or controlled recordings,
limiting scene diversity despite large video counts (\eg 3M videos
from only 70 scenes in PhyWorld). YoCausal incorporates \emph{any}
real-world video at zero cost and is extensible, achieving the highest scene diversity among all listed benchmarks. $\uparrow$ indicates that video count and scene coverage grow continuously as new subsets are added.  (2D): simulation in two-dimensional. (Laboratory): videos recorded under controlled laboratory setups.
}
\setlength{\tabcolsep}{14pt}  
      \resizebox{0.75\linewidth}{!}
    {%
      \begin{tabular}{@{}lccc@{}}
      \toprule
      & Video type & \# Video & \# Video scene  \\
      \midrule
      PhyWorld~\cite{kang2024far}
      & Synthetic (2D) & \textbf{3M}  & 70 \\
      LikePhys~\cite{yuan2025likephys}
      & Synthetic & 120 & 12\\
      Physion~\cite{bear2021physion}
      & Synthetic & 10400 & 260 \\
      IntPhys2~\cite{bordes2025intphys}
      & Synthetic & 1416 & 344\\
      Phys101~\cite{wu2016physics}
      & Real-World (Laboratory) & 2500 & 101 \\
      Physics IQ~\cite{motamed2026generative}
      & Real-World (Laboratory) & 396 & 132\\\hdashline[0.5pt/2pt]
      Ours
      & Real-World & 1232$\uparrow$ & \textbf{1232$\uparrow$}\\

      \bottomrule
      \end{tabular}%
    }%
\label{tab:dataset}
\end{table*}

Fig.~\ref{fig:overview} overviews our framework. We describe the \emph{extensible dataset} construction (Sec.~\ref{sec:dataset}), formalize the link between a diffusion model's \emph{``surprise''} and its \emph{denoising loss} (Sec.~\ref{sec:surprise}), then introduce the \emph{Reverse Surprise Index} (RSI) for arrow-of-time perception (Sec.~\ref{sec:rsi}) and the \emph{Causality Cognition Index} (CCI) for disentangling genuine causal understanding (Sec.~\ref{sec:cci}).

\subsection{Dataset Construction}
\label{sec:dataset}
As mentioned in \cref{sec:intro}, we use reversed video to validate models' causal cognition ability. Our benchmark can utilize \emph{any} real-world video at zero cost. This enables building a benchmark of arbitrary scale and scene diversity without synthetic rendering or controlled setups.

\begin{figure*}[t]
    \centering
    \includegraphics[width=1.0\linewidth]{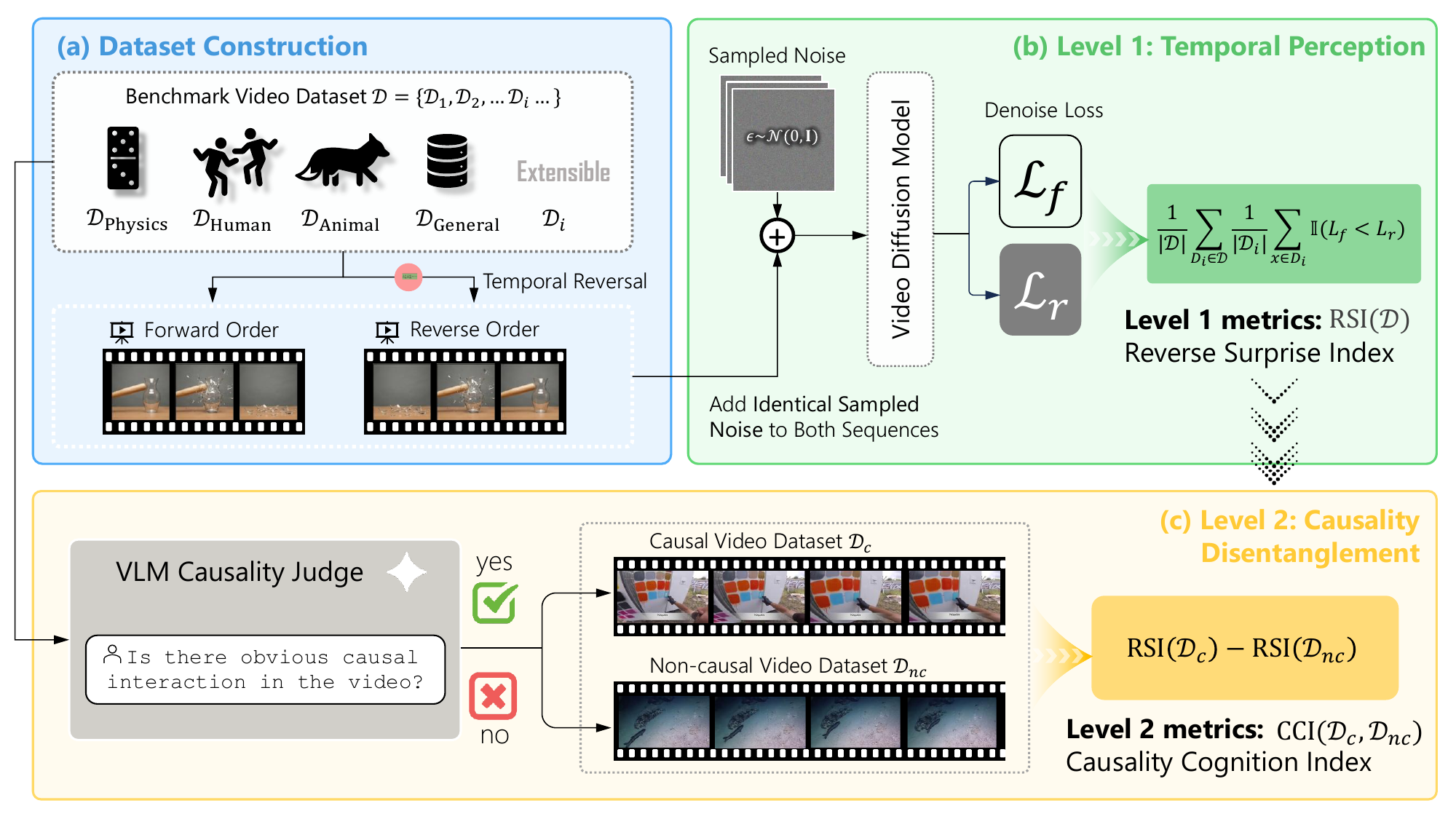}
    \caption{
    \textbf{Overview of the YoCausal evaluation framework.}
    \textbf{(a) Dataset Construction:} We construct an infinitely extensible benchmark by using real-world videos from different domains.
    By applying zero-cost temporal reversal, we generate natural counterfactual pairs (forward $x^f$ and reverse $x^r$).
    \textbf{(b) Level 1 Temporal Perception:} Identical sampled noise $\epsilon$ is added  to both sequences and compute their denoising losses.
    Reverse Surprise Index (RSI), quantifies the model's perception of the arrow of time by measuring the proportion of instances where the reversed video has a higher loss ($\mathcal{L}_r > \mathcal{L}_f$).
    \textbf{(c) Level 2 Causality Disentanglement:} To disentangle genuine causal cognition from statistical temporal biases, a Vision-Language Model (VLM) divides the dataset into causal ($\mathcal{D}_c$) and non-causal ($\mathcal{D}_{nc}$) subsets.
    The Level-2 metric, Causality Cognition Index (CCI), is computed as the difference in RSI between these subsets, isolating the model's genuine causal cognition ability.
    }
    \label{fig:overview}
\end{figure*}

As shown in ~\cref{fig:overview}(a), we construct a dataset $\mathcal{D} = \{\mathcal{D}_1, \mathcal{D}_2, \dots, \mathcal{D}_i, \dots\}$ of thematic subsets. Unlike closed-form benchmarks, our design is arbitrarily extensible: new subsets can be seamlessly added. In this paper, we use four representative subsets of everyday scenes: \textbf{General} $\mathcal{D}_{General}$ (unconstrained daily-life events), \textbf{Physics} $\mathcal{D}_{Physics}$ (mechanics, optics, thermodynamics, \etc), \textbf{Human Action} $\mathcal{D}_{Human}$ (diverse human activities), and \textbf{Animal Action} $\mathcal{D}_{Animal}$ (various animal behaviors), sourced from existing datasets: Moment in Time~\citep{monfort2019moments}, Physics IQ\citep{motamed2026generative}, Kinetics\citep{kay2017kinetics} and Animal Kingdom~\citep{ng2022animal} (details are provided in \cref{suppl_sec:dataset}). Performance breakdowns across subsets reveal each model's domain-specific strengths and weaknesses. Notably, future researchers can integrate additional domains (\eg, tool use) to keep the benchmark evolving alongside model capabilities.

As summarized in ~\cref{tab:dataset}, prior physics benchmarks rely on synthetic data or small-scale controlled recordings, whereas our method incorporates any real-world videos at zero cost. Consequently, YoCausal achieves significant breakthroughs in scale and real-world scene coverage.

\subsection{Formulating Surprise via Denoising Loss}
\label{sec:surprise}

We quantify a VDM's surprise at a video by its denoising loss. Following \emph{violation of expectation} (VoE) paradigm from cognitive science~\citep{leslie1987six}, where the degree of surprise at expectation-violating events reveals whether an internal model has formed. We treat the learned distribution as the model's internal expectation: lower assigned probability means greater surprise. This framing allows us to probe a VDM's cognitive priors including causal understanding.

Concretely, a VDM learns the data distribution $p(x)$ of videos by training a neural network $\epsilon_\theta$ to denoise noisy inputs $x_t$. Its denoising loss is formulated as~\cref{eq:denoise_loss_upper_bound}:
%
{
\small
\begin{equation}
    \mathcal{L}_{\text{denoise}}(\theta;x_t) = \mathbb{E}_{t \sim \mathcal{U}(1,T),\, \epsilon \sim \mathcal{N}(0,\mathbf{I})} \!\left[ \left\| \epsilon - \epsilon_\theta(x_t, t) \right\|_2^2 \right] \gtrsim \mathbb{E}_{x_0}[-\log p_\theta(x_0)].
    \label{eq:denoise_loss_upper_bound}
\end{equation}
}
%

By variational inference theory~\citep{ho2020denoising,sohl2015deep}, this loss upper-bounds the negative log-likelihood (NLL). This means that the denoising loss can serve as an empirical proxy for the NLL of the video sequence $x_0$ within the model's learned distribution.
Specifically, a higher denoising loss thus indicates lower model-assigned probability, letting us translate ``degree of surprise'' into magnitude of denoising loss as a quantifiable metric.
The validity of denoising loss as a likelihood proxy has been established from prior work~\citep{li2023your,clark2023text,yuan2025likephys,song2020score,song2021maximum,kingma2021variational}.

\subsection{Level 1: Measuring Arrow-of-Time Perception via RSI}
\label{sec:rsi}

Our first-level metric asks a simple question: is a model more surprised by a video played backward? 
Following infant studies~\citep{leslie1987six}, a model that has internalized the arrow of time and causality, like an infant, should assign lower likelihood to the reversed video $x^r$ than to the forward one $x^f$. Building on the surprise signal in~\cref{sec:surprise}, we can express $p_\theta(x^f)>p_\theta(x^r)$ equivalently as $\mathcal{L}_{\text{denoise}}(\theta; x^r) > \mathcal{L}_{\text{denoise}}(\theta; x^f)$.



\subsubsection{Reverse Surprise Index.}
We propose the \emph{Reverse Surprise Index} (RSI~\cref{eq:rsi}) as our Level-1 metric.
For each video $x_{i,j}\!\in\!\mathcal{D}_i$, let $x^f_{i,j}$ and $x^r_{i,j}$ denote its forward and reversed sequences, respectively.
We uniformly sample $K{=}10$ timesteps from the diffusion process and apply identical Gaussian noise $\boldsymbol{\epsilon}$ to both sequences at each timestep (\cref{fig:overview}(b)), then average the resulting losses.
Fixing the timesteps and noise ensures identical denoising difficulty. More details are provided in \cref{suppl_sec:RSI}. Concretely, RSI measures the proportion of videos for which the model correctly assigns a lower denoising loss to the forward sequence. For a dataset $\mathcal{D}$ composed of sub-datasets $\{\mathcal{D}_i\}$, we compute the average across subset:
{
\small
\begin{equation}
  \mathrm{RSI}(\mathcal{D})
  = \frac{1}{|\mathcal{D}|}
    \sum_{\mathcal{D}_i \in \mathcal{D}}
    \frac{1}{|\mathcal{D}_i|}
    \sum_{x_{i,j} \in \mathcal{D}_i}
    \mathbbm{1}\!\bigl[
      \mathcal{L}_{\text{denoise}}(\theta;\,x^r_{i,j})
      >
      \mathcal{L}_{\text{denoise}}(\theta;\,x^f_{i,j})
    \bigr],
  \label{eq:rsi}
\end{equation}
}
where $\mathbbm{1}[\cdot]$ is the indicator function and $\mathrm{RSI}\!\in\![0,1]$; higher values indicate stronger perception of the arrow of time and causality.

\subsection{Level 2: Disentangling Causality via CCI}
\label{sec:cci}

We introduce the Causality Cognition Index (CCI) to disentangle a model's understanding of causality from its sensitivity to temporal direction.

\begin{figure*}[t]
    \centering
    \includegraphics[width=1.0\linewidth]{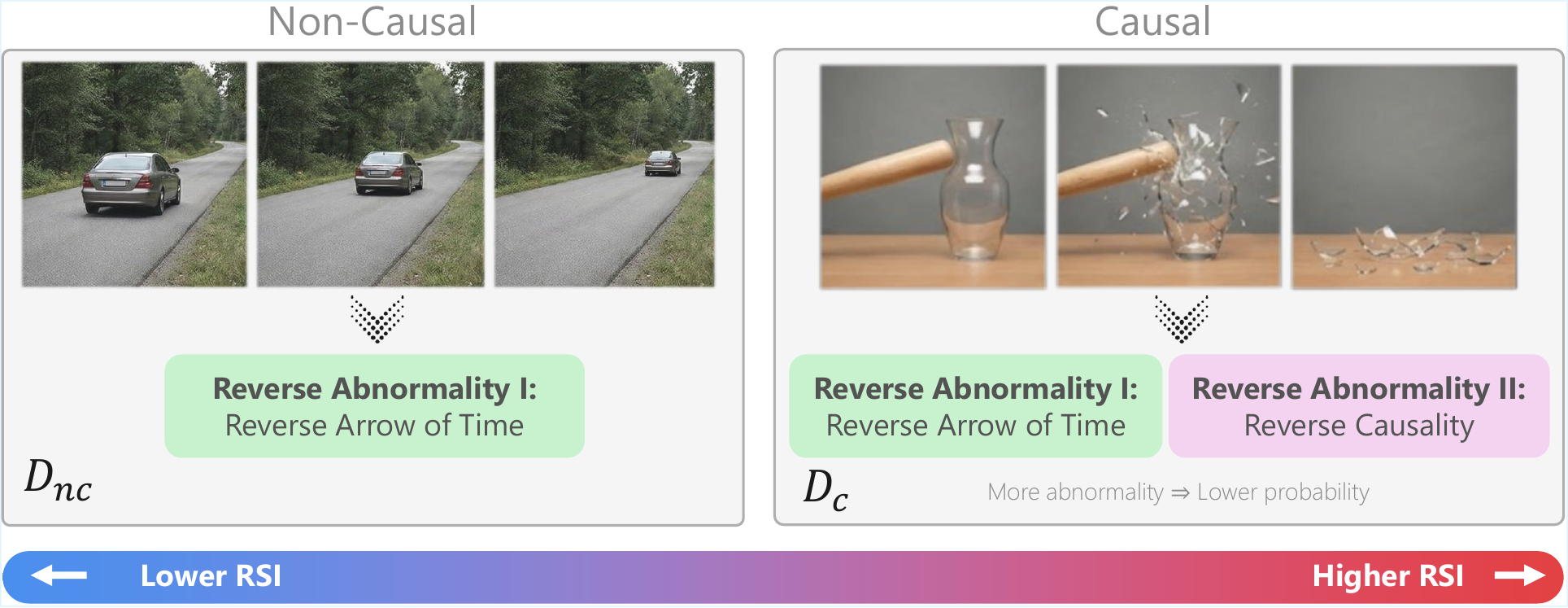}
    \caption{
    \textbf{Key idea behind CCI.}
    Reversing a non-causal video (\eg a car cruising) introduces only
    one source of abnormality: reversed arrow of time. While, 
    a causal video (\eg a hammer shattering a vase) introduces an
    additional source: reversed causality. A causally aware model should
    perceive more abnormality in the reversed causal video, exhibiting
    greater surprise (higher RSI). The gap
    $\mathrm{CCI}=\mathrm{RSI}(\mathcal{D}_c)
    -\mathrm{RSI}(\mathcal{D}_{nc})$ thus disentangles the model's
    sensitivity to causality from its sensitivity to statistical
    temporal patterns.
    }
    \label{fig:CCI_idea}
\end{figure*}

\textbf{Blind spot of RSI.}
RSI cannot reveal why a model is surprised: reversal flips both the arrow of time and causality. Leslie and Keeble\cite{leslie1987six} resolved this ambiguity in infants using hand-crafted causal and non-causal stimuli (\cref{fig:CCI_idea}), showing that infants were more surprised by reversed causal videos, which implies that they perceive causality.

\subsubsection{Causality Cognition Index.}

Following Leslie and Keeble~\cite{leslie1987six}, we need causal and non-causal videos. Real-world videos naturally vary in \emph{causal salience} (\cref{fig:CCI_idea}): breaking glass exhibits a clear causal chain, while a car cruising on a highway does not. Therefore, there is no need to craft synthetic videos, and we directly partition $\mathcal{D}$ into a causal subset $\mathcal{D}_c$ and a non-causal subset $\mathcal{D}_{nc}$ based on whether cause-effect interactions are present.

A causally aware model should therefore show higher RSI on $\mathcal{D}_c$ than on $\mathcal{D}_{nc}$, and we define CCI as~\cref{eq:cci}. A higher CCI indicates a model captures reversed causality cues.
{
\small
\begin{equation}
    \text{CCI}(\mathcal{D}) = \text{RSI}(\mathcal{D}_c) - \text{RSI}(\mathcal{D}_{nc}).
    \label{eq:cci}
\end{equation}
}
%



As shown in Fig.~\ref{fig:overview}(c), since constructing CCI only requires detecting whether causality exists, which is easier than judging its correctness, we automate dataset splitting with an advanced Vision-Language Model (VLM) using a carefully designed prompt (see \cref{suppl_sec:CCI}) to ensure scalability. We validate the reliability of this split in \cref{sec:reliability}.

Finally, we must emphasize that CCI is a relative index and must be jointly interpreted with RSI: high RSI but low CCI suggests the model only perceives the statistical arrow of time; high CCI but low RSI renders the CCI unreliable due to poor temporal grounding. We introduce an aggregate ranking combining both metrics in Sec.~\ref{sec:aggregate_rank}.

\section{Experiment}
\label{sec:exp}


We employ YoCausal to evaluate the causal cognition of current
open-source video diffusion models, presenting Level~1 RSI results
(\cref{sec:rsi_result}), Level~2 CCI results (\cref{sec:cci_result}),
an aggregate ranking (\cref{sec:aggregate_rank}) that jointly considers both metrics and cross-metric analysis (\cref{sec:cross_metric}). 

\paragraph{Settings.}
We evaluate 13 state-of-the-art open-source text-to-video diffusion models spanning diverse architectures and scales~\citep{guo2023animatediff,hong2022cogvideo, yang2024cogvideox,genmo2024mochi,kong2024hunyuanvideo,wan2025wan,hacohen2024ltx} (details in \cref{suppl_sec:model_setting}). To ensure each model operates at its best, all inference configurations strictly follow official recommendations, including FPS, resolutions and so on. Videos are preprocessed through reasonable resizing and temporal adjustments to match each model's specifications. Detailed per-model settings and preprocessing procedures are provided in \cref{suppl_sec:model_setting,suppl_sec:video_preprocessing}.

\subsection{RSI Results}
\label{sec:rsi_result}
\begin{figure*}[t]
    \centering
    \includegraphics[width=1.0\linewidth]{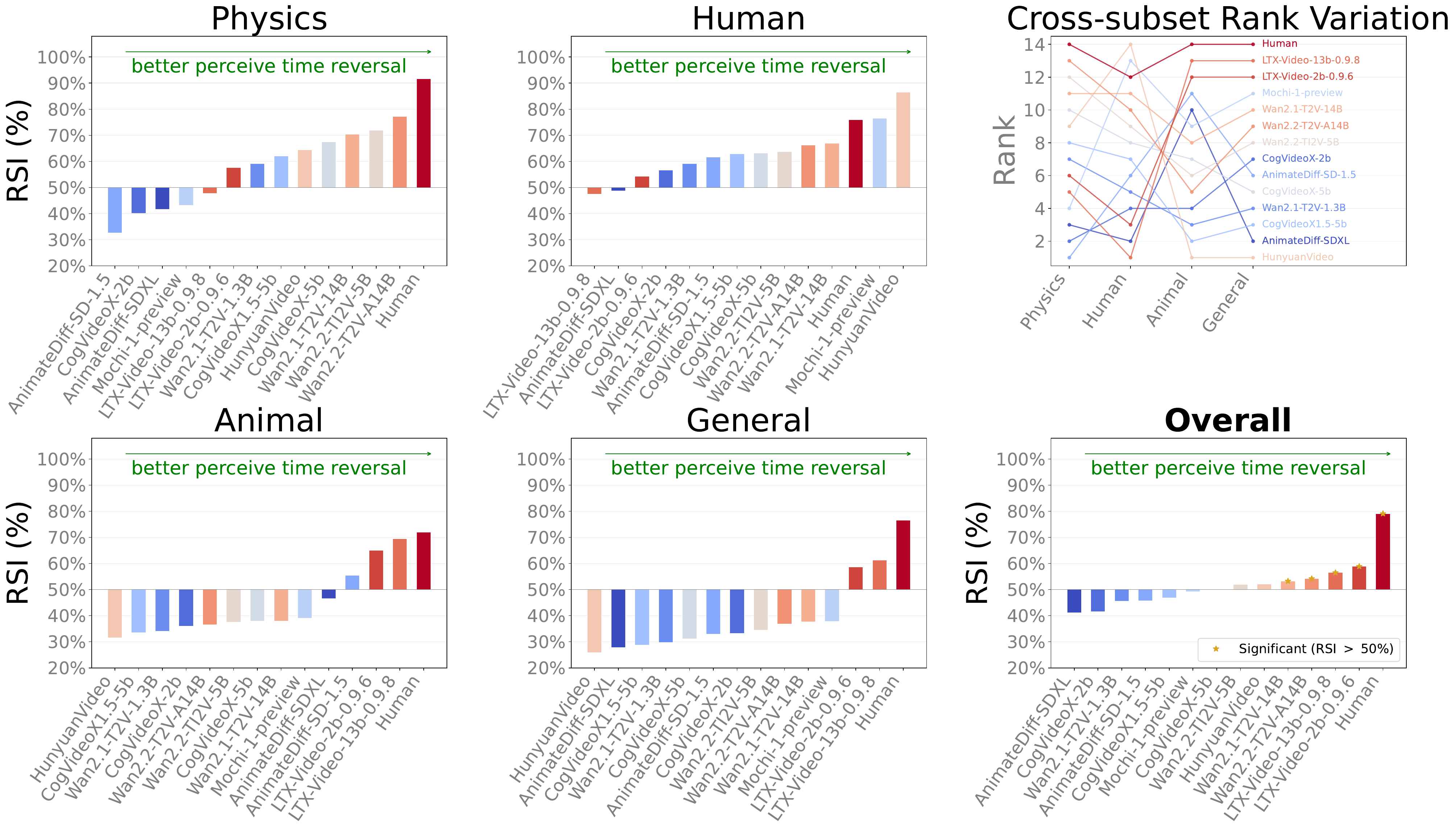}
    \caption{
    \textbf{Level-1 RSI results.}
    RSI scores of 13 VDMs and human annotators across four subsets
    and the full dataset. Several models surpass the 50\% random-guess
    baseline with 90\% confidence (bootstrap test) but still lag considerably behind human performance.
    Cross-subset variations highlight the necessity of evaluating
    across diverse domains.
    }
    \label{fig:RSI}
\end{figure*}

In \cref{fig:RSI}, we evaluate all 13 models on Level~1 RSI, with
human annotators judging 1{,}232 videos' temporal directions as reference~\citep{hanyu2023ready} (details
in \cref{suppl_sec:human_annotating}). Humans achieve the highest RSI across
all subsets except Human Action. Several models surpass the 50\%
random-guess baseline with 90\% confidence (bootstrap test), yet a significant gap remains relative to
human performance. Higher-fidelity models (\eg LTX-Video-13B, Wan2.1/2.2-14B) tend to score higher,
whereas lower-fidelity models (\eg AnimateDiff-SDv1.5/SDXL) exhibit weaker temporal perception.


Per-subset results reveal cross-domain variation due to (1)~differing
cue strength across subsets: for example, videos in $D_{Physics}$ contain unambiguous
anomalies when reversed, and (2)~domain-specific biases from training data: for instance, most models perform well on $D_{Human}$ given the abundance of human activity videos online. These highlight the value of evaluating across diverse subsets.


Notably, some models score below the 50\%, suggesting their learned distributions capture local visual smoothness without internalizing the arrow of time and yield no preference, or even a slight inverse preference for forward sequences. A similar observation is reported in LikePhys~\citep{yuan2025likephys}.



\subsection{CCI Results}
\label{sec:cci_result}

\begin{figure*}[t]
    \centering
    \includegraphics[width=1.0\linewidth]{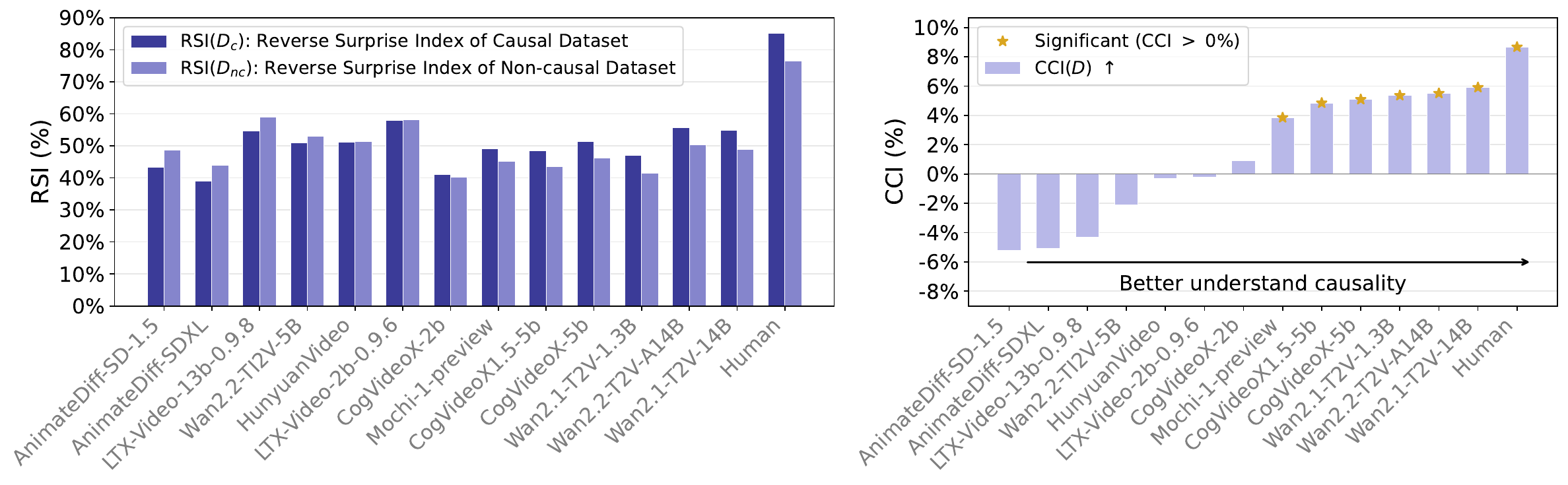}
    \caption{
    \textbf{Level-2 CCI results.}
    Left: RSI scores of 13 VDMs and human annotators on the causal
    subset $\mathcal{D}_c$ (dark blue) and non-causal subset
    $\mathcal{D}_{nc}$ (light blue). Right: the resulting CCI. 
    }
    \label{fig:CCI}
\end{figure*}


We further analyze Level~2 CCI in \cref{fig:CCI}. Humans achieve the highest CCI, though the margin over models is modest since humans already saturate both subsets. Several models attain positive CCI with 90\% confidence (bootstrap), demonstrating preliminary causal perception; the top performers concentrate in the Wan and CogVideo families, hinting that shared training data and architectural choices may give rise to emergent causal understanding. Crucially, models ranking high on RSI—LTX-Video-2B/13B and HunyuanVideo—score poorly on CCI, confirming that our two-level framework disentangles causal cognition from mere arrow-of-time perception. 

Negative CCI in some models reflects the same distributional deficiency noted in \cref{sec:rsi_result}: lacking an internalized causality, they treat reversed causal and non-causal sequences as equally anomalous.



\subsection{Aggregating Arrow of Time and Causality Cognition}
\label{sec:aggregate_rank}
\begin{figure*}[t]
    \centering
    \begin{minipage}[c]{0.45\linewidth}
        \includegraphics[width=\linewidth]{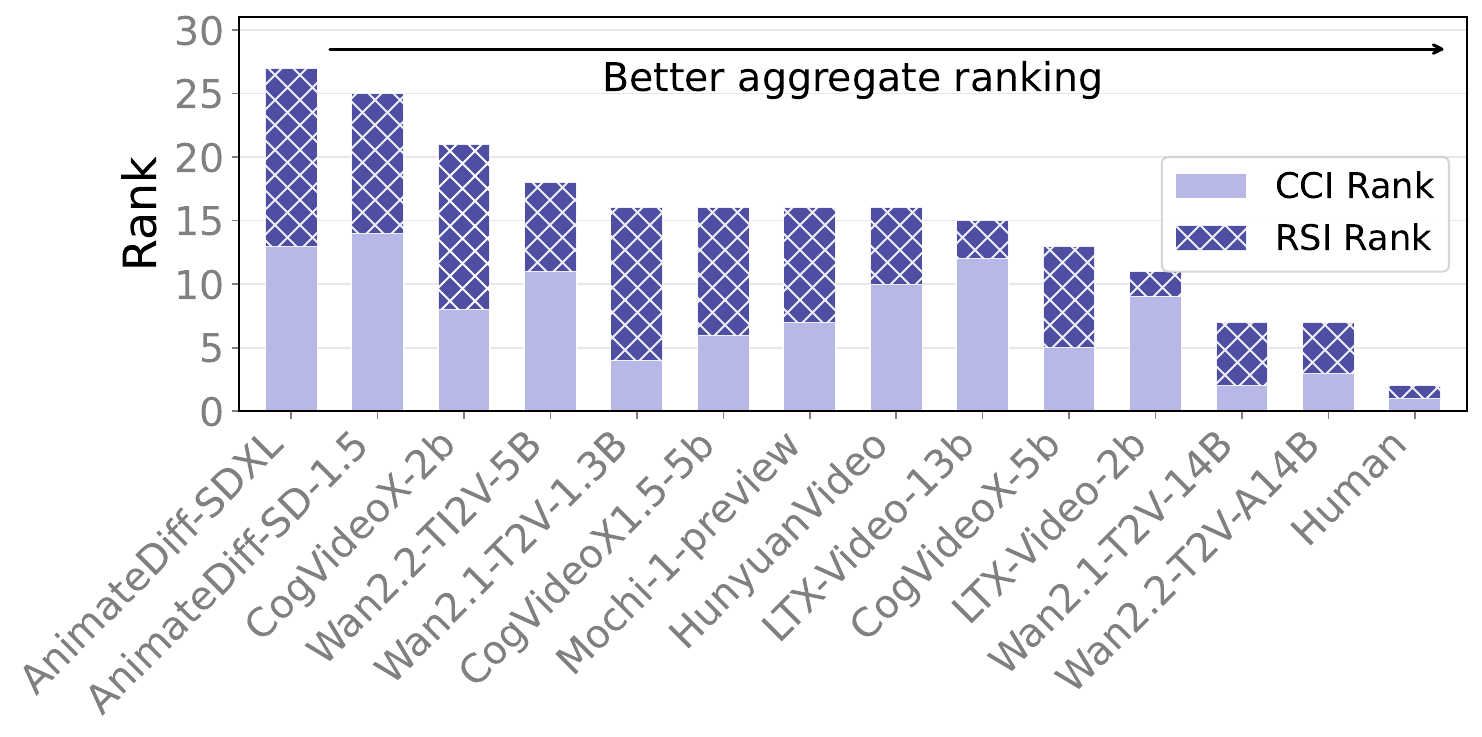}
    \end{minipage}%
    \hspace{0.02\linewidth}%
    \begin{minipage}[c]{0.52\linewidth}
        \caption{
        \textbf{Aggregate ranking of causal cognition.}
        We sum each model's RSI and CCI ranks to obtain an aggregate causality score (lower is better). Ties are broken by RSI rank.
        }
        \label{fig:aggregat_rank}
    \end{minipage}
\end{figure*}


As discussed in \cref{sec:cci}, RSI and CCI should not be interpreted in isolation, and robust causal understanding requires strong performance on both metrics. To provide a single, intuitive measure of overall causal cognition, we combine the two indices with direct arithmetic combination. We propose a heuristic aggregate rank by summing each model's ranks on RSI and CCI as aggregate causality score and sort the totals. Since RSI serves as the prerequisite foundation for causal cognition, ties are broken in favor of the model with the higher RSI rank.

As shown in \cref{fig:aggregat_rank}, this aggregate ranking provides a holistic view of each model's causal cognition capability and will serve as the quantitative basis for the cross-metric analyses (\cref{sec:cross_metric}).

\subsection{Cross-Metric Analysis}
\label{sec:cross_metric}




\begin{table*}[t]
\small
\setlength{\tabcolsep}{3pt}
\centering
\caption{
\textbf{Cross-metric analysis.}
Kendall's $\tau$ between our aggregate causal cognition ranking and external benchmarks/model features. High correlation with human preference validates our benchmark; positive correlation with \citep{yuan2025likephys} indicates causal cognition relates to physical understanding; low correlation with aesthetic quality rules out visual-appeal confounding; strong correlation with release date and parameters supports scaling laws and architectural evolution.
}
\setlength{\tabcolsep}{8pt}
\resizebox{\textwidth}{!}{%
  \begin{tabular}{@{}lccccccccc@{}}
  \toprule
   & 
   & 
   & \multicolumn{5}{c}{VBench~\citep{huang2024vbench}} \\
   \cmidrule(l){4-8}
   & \shortstack{Human\\Preference} 
   & \shortstack{LikePhys\\\citep{yuan2025likephys}}
   & \shortstack{Aesthetic\\Quality}
   & \shortstack{Subject\\Consistency}
   & \shortstack{Background\\Consistency}
   & \shortstack{Motion\\Smoothness}
   & \shortstack{Temporal\\Flickering}
   & \shortstack{Release\\Date}
   & \shortstack{\# of\\Parameters } \\
  \midrule
  Kendall's $\tau$
   & 0.6667 & 0.5111 & 0.1026 & 0.3846 & 0.1282 & 0.2308 & 0.1538 & 0.4576 & 0.6584 \\
  p-value
   & 0.0009 & 0.0466 & 0.6754 & 0.0763 & 0.5900 & 0.3062 & 0.5098 & 0.0318 &  0.0024 \\
  \bottomrule
  \end{tabular}%
}%
\label{tab:cross_metrics}
\end{table*}

To validate our benchmark and explore the interplay between causality and other model capabilities, we compute Kendall's rank correlation coefficient $\tau$\citep{abdi2007kendall} between our aggregate rank and model rankings on several external metrics and model properties, as summarized in \cref{tab:cross_metrics}.

\paragraph{Human Causality Preference.}
To verify that our benchmark captures causal understanding, we conduct a user study in which models generate videos from causally rich prompts and human evaluators compare them by causality plausibility (details in \cref{suppl_sec:human_preference}). As shown in \cref{tab:cross_metrics}, our benchmark exhibits a high correlation ($\tau$ = 0.6667) with human preference.


\paragraph{Intuitive Physics Correlation.}
Clarifying the relationship between causal cognition and intuitive physics will provide important guidance for future model improvement. We compute the rank correlation with LikePhys~\citep{yuan2025likephys}, a benchmark for intuitive physics in VDMs. \cref{tab:cross_metrics} reveals positive correlation ($\tau = 0.5111$), implying that physical laws constrain object interactions and thereby influence causality. However, the magnitude shows causal cognition is not reducible to physical intuition alone, meaning that separately assessing causal capabilities remains essential, highlighting the unique contribution of our benchmark.

\paragraph{VBench Correlation.}
 In addition, we examine the relationship between causality and visual generation quality by computing correlations with five VBench metrics~\citep{huang2024vbench}. As shown in \cref{tab:cross_metrics}, causality shows low correlation with aesthetic quality ($\tau = 0.1026$), suggesting that our benchmark is not confounded by visual appeal. The subject consistency index shows mild correlations: Inconsistency violates causal expectations.

\paragraph{Scaling Law and Generation Evolution.}
As shown in \cref{tab:cross_metrics}, release date ($\tau{=}0.4576$) and model parameters ($\tau{=}0.6584$) correlate significantly with causality cognition, implying scaling laws~\citep{yin2025towards,kaplan2020scaling} and architectural evolution improve causal cognition (\cref{suppl_sec:scaling_law}).

\section{Benchmark Reliability and Validity}
\label{sec:reliability}

\begin{figure*}[t]
    \centering
    \begin{subfigure}[t]{0.44\linewidth}
        \centering
        \includegraphics[height=3.5cm]{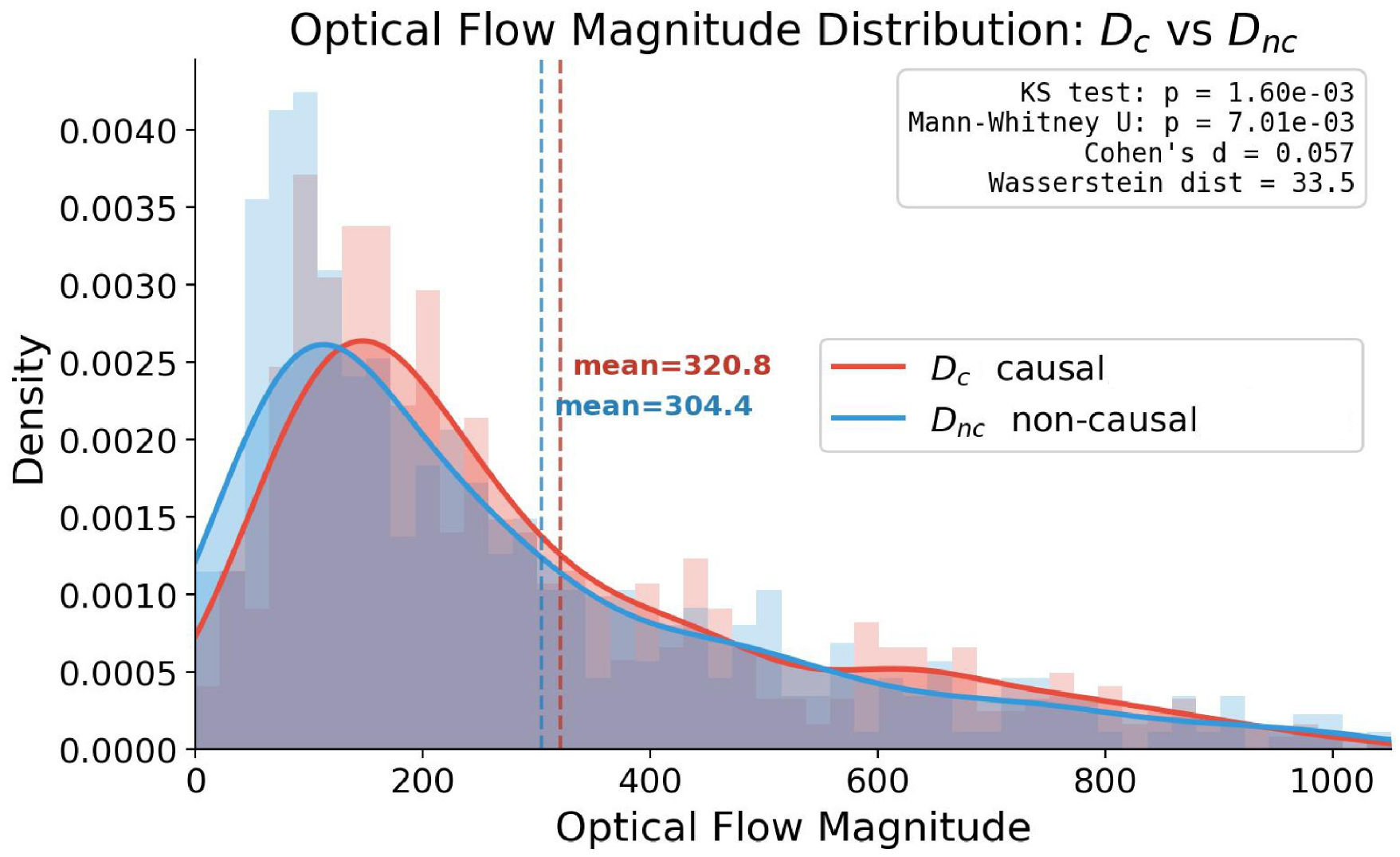}
        \caption{{Optical flow magnitude distributions of $\mathcal{D}_c$ and $\mathcal{D}_{nc}$ (computed with RAFT~\cite{teed2020raft}).}}
        \label{fig:dc_dnc_distribution}
    \end{subfigure}
    \hfill
    \begin{subfigure}[t]{0.44\linewidth}
        \centering
        \includegraphics[height=3.5cm]{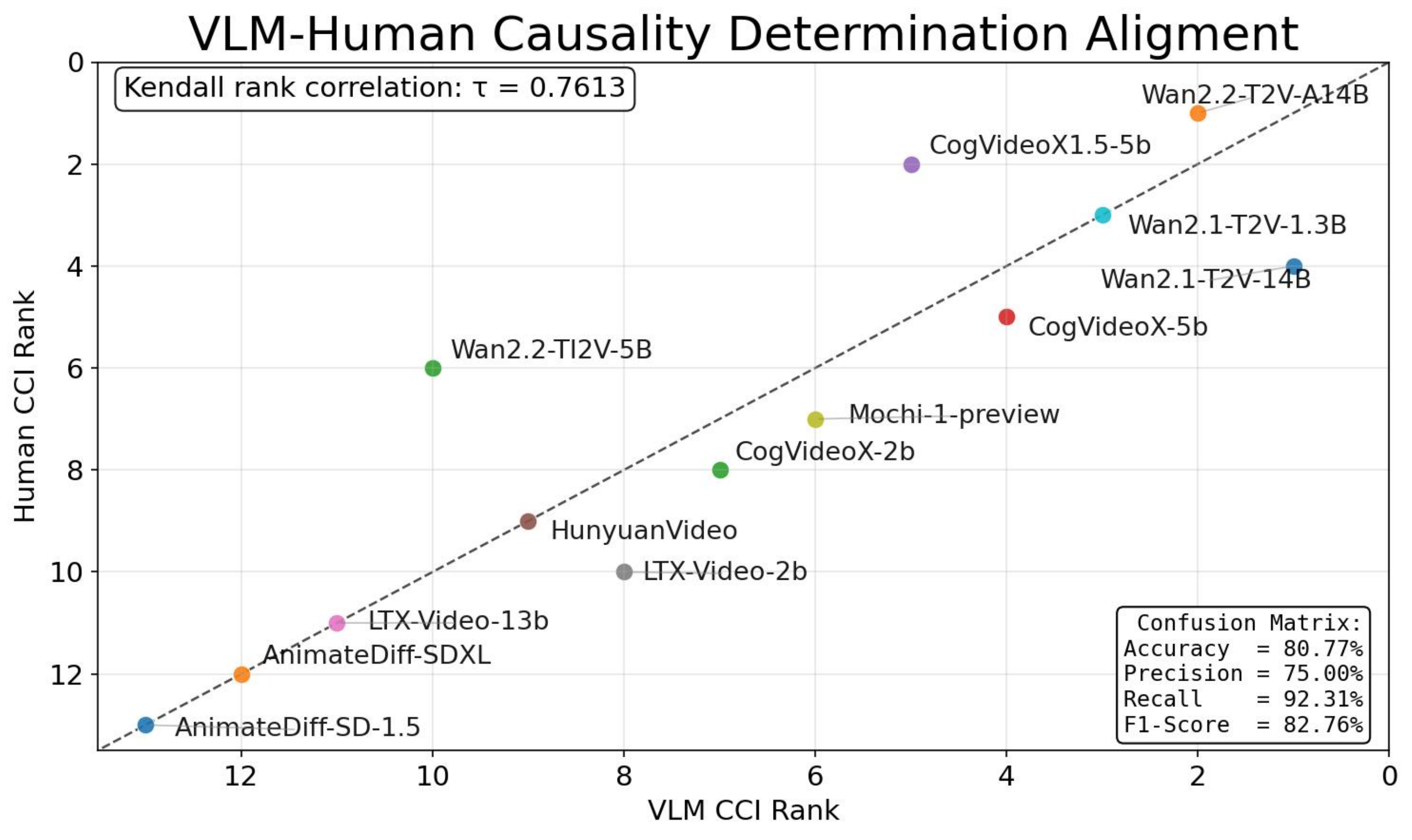}
        \caption{{VLM-human Alignment.}}
        \label{fig:VLM-human-align}
    \end{subfigure}
    \caption{\textbf{Validating VLM-based causal/non-causal partitioning.} (a) The effect size between two optical flow distribution is negligible (Cohen's $d{=}0.057{<}0.2$), confirming the VLM infers causality from high-level semantics rather than low-level motion cues. (b) Comparing CCI rankings and the confusion matrix from a VLM partition over the full dataset against human annotators (30 clearly causal, 30 clearly non-causal videos), the high Kendall correlation ($\tau{=}0.7613$) and F1-score (82.76\%) confirm VLM-based labeling as a reliable proxy for human judgment.}
    \label{fig:vlm_validation}
\end{figure*}



\begin{table*}[h]
\small
\setlength{\tabcolsep}{3pt}
\centering
\caption{
\textbf{VLM sensitivity of aggregate rankings.} Kendall's $\tau$ between
aggregate rankings induced by three different VLMs. The consistently high correlations confirm that our benchmark is robust to the choice of VLM judge.
}
\setlength{\tabcolsep}{8pt}
\resizebox{0.95\textwidth}{!}{%
  \begin{tabular}{@{}lccccccc@{}}
  \toprule
    Kendall's $\tau$ / p-value
   & \shortstack{Gemini 3.0 Pro~\cite{team2023gemini}} 
   & \shortstack{GPT-4o~\cite{hurst2024gpt}}
   & \shortstack{Qwen 3.5 9B~\cite{team2026qwen3,yang2025qwen3}} \\
  \midrule
  Gemini 3.0 Pro
   & 1.000 / 0.0000 & 0.6923 / 0.0005 & 0.6666 / 0.0009 \\
  GPT-4o
   & 0.6923 / 0.0005 & 1.000 / 0.0000 & 0.6666 / 0.0009 \\
   Qwen 3.5 9B
   & 0.6666 / 0.0009 & 0.6666 / 0.0009 & 1.000 / 0.0000 \\
  \bottomrule
  \end{tabular}%
}%
\label{tab:VLM_sensitivity}
\end{table*}


\subsection{VLM Reliability}
\label{sec:vlm_reliability}
CCI relies on a VLM, Gemini~3.0~Pro~\citep{team2023gemini} by default, to partition the dataset (Sec.~\ref{sec:cci}), so its validity depends on the reliability of VLMs.
We validate it from three aspects: agreement with human annotation, independence from low-level motion cues, and robustness to the choice of VLMs.

\textbf{Human Alignment.}
VLM labels agree well with human annotation (F1 = 82.76\%), and CCI-based model rankings from VLM and human labels are highly consistent (Kendall's $\tau=0.761$, Fig.~\ref{fig:vlm_validation}b).
Moreover, $\mathcal{D}_c$ and $\mathcal{D}_{nc}$ show negligible difference in optical-flow magnitude (Cohen's $d=0.057$, Fig.~\ref{fig:vlm_validation}a), indicating the VLM partitions videos by causality rather than low-level motion cues.

\textbf{VLM Sensitivity.}
Re-partitioning with GPT-4o~\citep{hurst2024gpt} and the open-weight Qwen3.5-9B~\citep{team2026qwen3,yang2025qwen3} yields aggregate rankings consistent with Gemini~3.0~Pro's ($\tau \in [0.667, 0.692]$, $p<0.001$; Table~\ref{tab:VLM_sensitivity}).
YoCausal is thus robust to the choice of VLM judge, and its partition should improve as VLMs advance.

\subsection{Confound Exclusion}
\label{sec:confound}
\textbf{Model-side.} For each video, the forward and reversed clips share the same model, preprocessing, timesteps, and noise, so these factors cancel in the pairwise comparison.
Since RSI is a win rate rather than a raw loss, it is also insensitive to model-specific loss scales and thus comparable across models; CCI inherits this property.

\textbf{Video-side.} Static factors (appearance, lighting) are identical across the pair and cancel out.
However, temporal factors (motion distribution and video smoothness), change under reversal.
CCI addresses this by differencing RSI across $\mathcal{D}_c$ and $\mathcal{D}_{nc}$: when the two subsets share similar temporal statistics, these factors cancel out.
As shown in Sec.~\ref{sec:vlm_reliability}, the two subsets have nearly identical motion-magnitude distributions, suggesting motion and smoothness does not account for the CCI.

\section{Conclusion and Limitation}
\label{sec:conclusion}

This paper presents YoCausal, the first benchmark for evaluating causal cognition in VDMs. Inspired by the VoE paradigm from cognitive science, we leverage temporally reversed real-world videos as natural counterfactual samples, establishing an arbitrarily extensible evaluation protocol free from synthetic data or controlled settings.
Experiments across 13 open-source VDMs reveal that: (1) perceiving the arrow of time is not equivalent to understanding causality; (2) even the best models exhibit a substantial gap relative to the human upper bound; (3) causal cognition is related to yet not reducible to intuitive physics and shows low correlation with aesthetic quality, confirming YoCausal captures a unique evaluation dimension, and (4) both parameter scaling and architectural evolution improve causal understanding. 
While physical understanding and scaling up models may partially improve causal perception~\citep{li2025pisa}, we hope YoCausal motivates the community~\citep{croitoru2023diffusion} to treat causal cognition as a distinct objective and advance this capability in the future.


\paragraph{Limitation.}
Our method struggles with temporally symmetric events (\eg Newton's cradle), where forward and reversed sequences are visually near-identical, rendering RSI ineffective.
Moreover, computing denoising losses requires access to model weights, limiting external evaluation of closed-source models, though model providers can still apply YoCausal internally to diagnose and improve causal cognition.
Addressing these limitations remains future work.


\clearpage

\crefalias{subsection}{appendix}

\section*{Appendix}
\renewcommand{\thesection}{A}
\renewcommand{\thesubsection}{\thesection.\arabic{subsection}}
\renewcommand{\thefigure}{\thesection.\arabic{figure}}
\renewcommand{\thetable}{\thesection.\arabic{table}}

\setcounter{figure}{0}
\setcounter{table}{0}



\section*{Appendix  Overview}
\label{suppl_sec:overview}

This appendix provides additional details and complete numerical results to complement the main manuscript. First, we describe the dataset construction protocol in \cref{suppl_sec:dataset}. Then, we detail the inference configurations and specifications of all 13 evaluated models in \cref{suppl_sec:model_setting}, followed by the unified video preprocessing pipeline in \cref{suppl_sec:video_preprocessing}. Next, we formalize the Reverse Surprise Index (RSI) and Causality Cognition Index (CCI) algorithms in \cref{suppl_sec:RSI} and \cref{suppl_sec:CCI}, respectively. We further discuss the potential prompt bias and how our two-level metric design addresses it in \cref{suppl_sec:prompt_bias}. Finally, we present the human annotation procedure and human preference study design in \cref{suppl_sec:human_annotating} and \cref{suppl_sec:human_preference}, and provide the complete numerical tabular results for all metrics in \cref{suppl_sec:numerical}.

\subsection{Details on Dataset Construction}
\label{suppl_sec:dataset}


\newcommand\QualitativeCompareBoxWidth{.18\textwidth}
\newcommand\QualitativeCompareImageWidth{.18\textwidth}

\begin{figure}[h]
    \centering
    \small
    \setlength{\tabcolsep}{0pt} 

    \resizebox{0.75\textwidth}{!}{%
\vspace{-1mm}
    \begin{tabular}{l@{\hspace{6pt}}c@{\hspace{-0.1pt}}c@{\hspace{-0.1pt}}c@{\hspace{-0.1pt}}c@{\hspace{-0.1pt}}c}

    \begin{tabular}{c} \rotatebox{90}{ \textsc{General}} \end{tabular}&%
    \parbox[c]{\QualitativeCompareBoxWidth}{\includegraphics[width=\QualitativeCompareImageWidth]{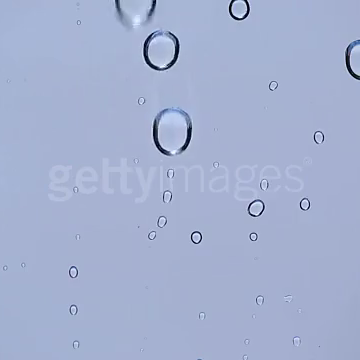}}&%
    \parbox[c]{\QualitativeCompareBoxWidth}{\includegraphics[width=\QualitativeCompareImageWidth]{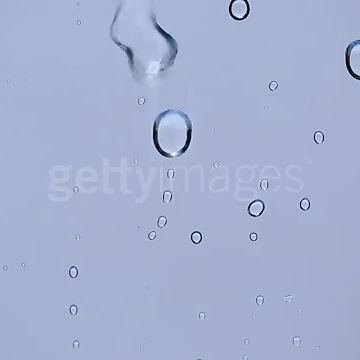}}&%
    \parbox[c]{\QualitativeCompareBoxWidth}{\includegraphics[width=\QualitativeCompareImageWidth]{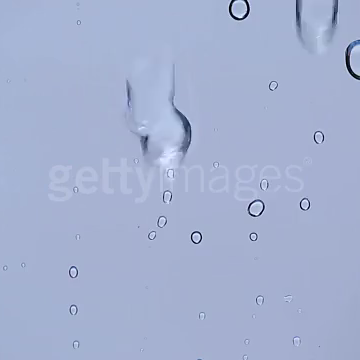}}&%
    \parbox[c]{\QualitativeCompareBoxWidth}{\includegraphics[width=\QualitativeCompareImageWidth]{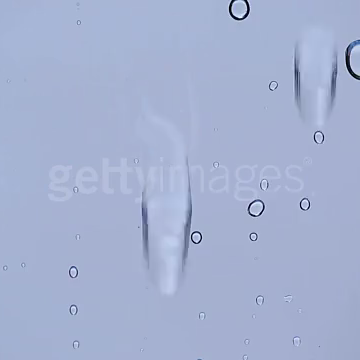}}&%
    \parbox[c]{\QualitativeCompareBoxWidth}{\includegraphics[width=\QualitativeCompareImageWidth]{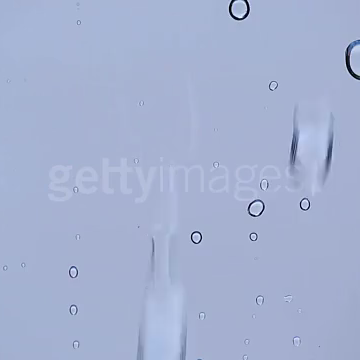}}%
    \\ \noalign{\vspace{2pt}} 

    \begin{tabular}{c} \rotatebox{90}{ \textsc{Physics}} \end{tabular}&%
    \parbox[c]{\QualitativeCompareBoxWidth}{\includegraphics[width=\QualitativeCompareImageWidth]{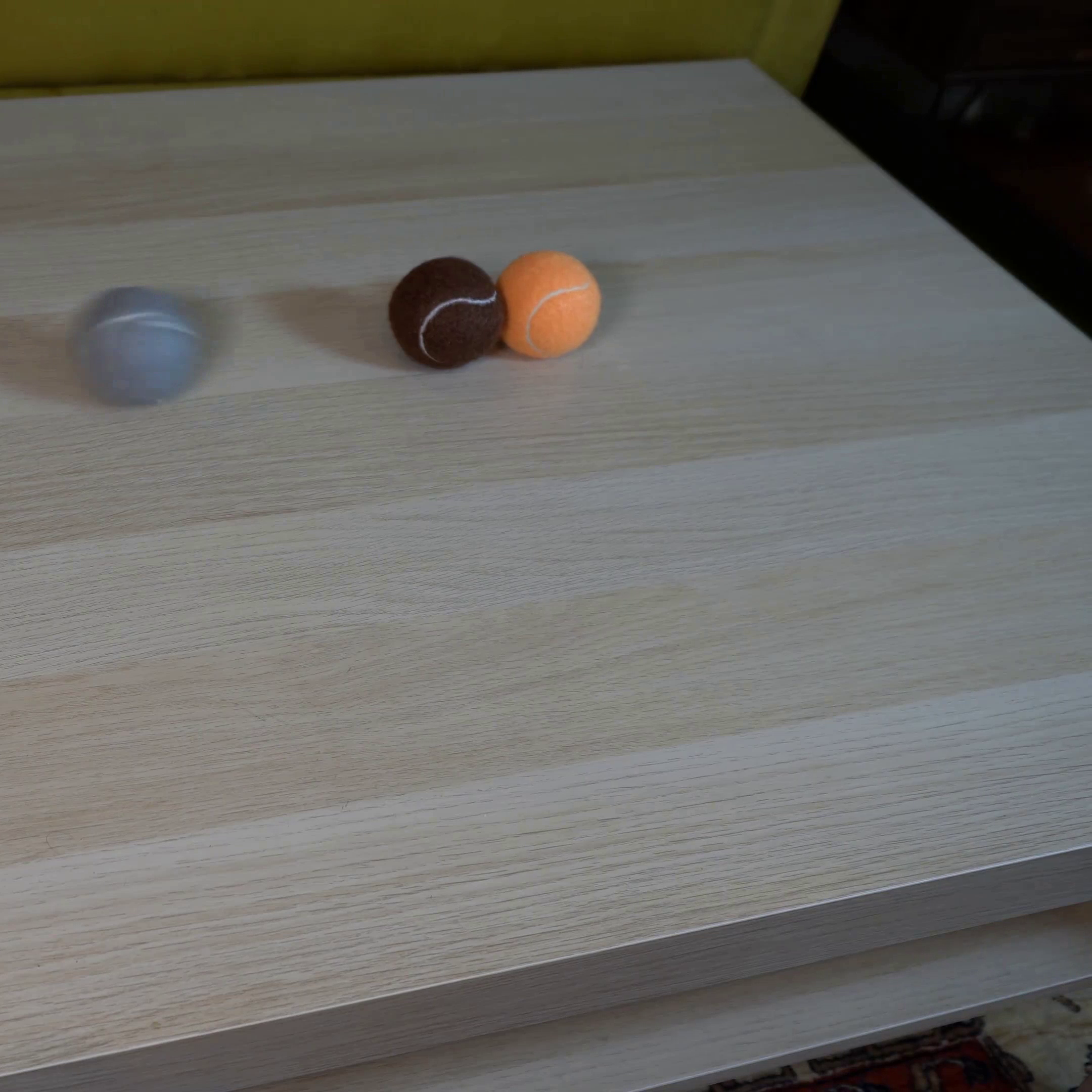}}&%
    \parbox[c]{\QualitativeCompareBoxWidth}{\includegraphics[width=\QualitativeCompareImageWidth]{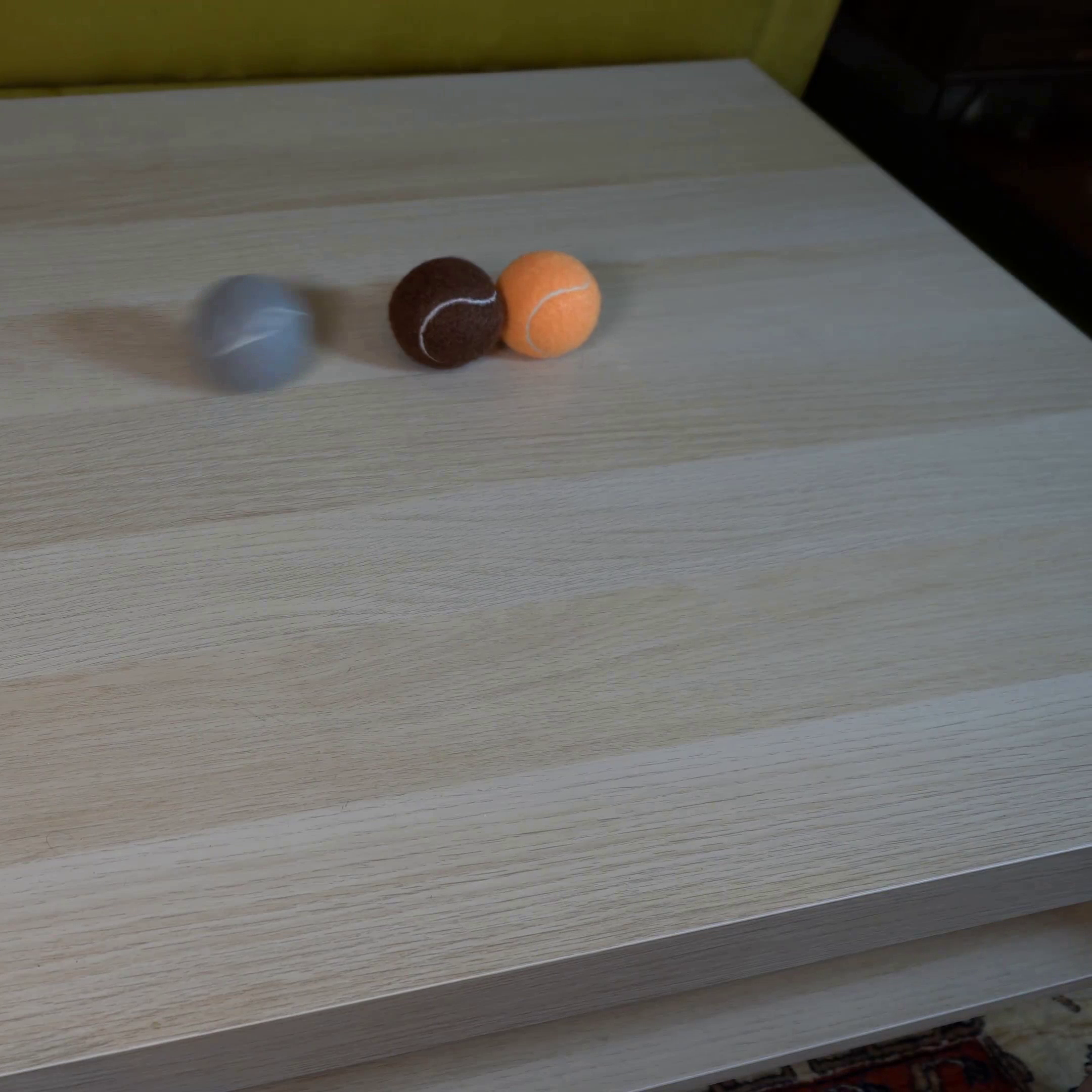}}&%
    \parbox[c]{\QualitativeCompareBoxWidth}{\includegraphics[width=\QualitativeCompareImageWidth]{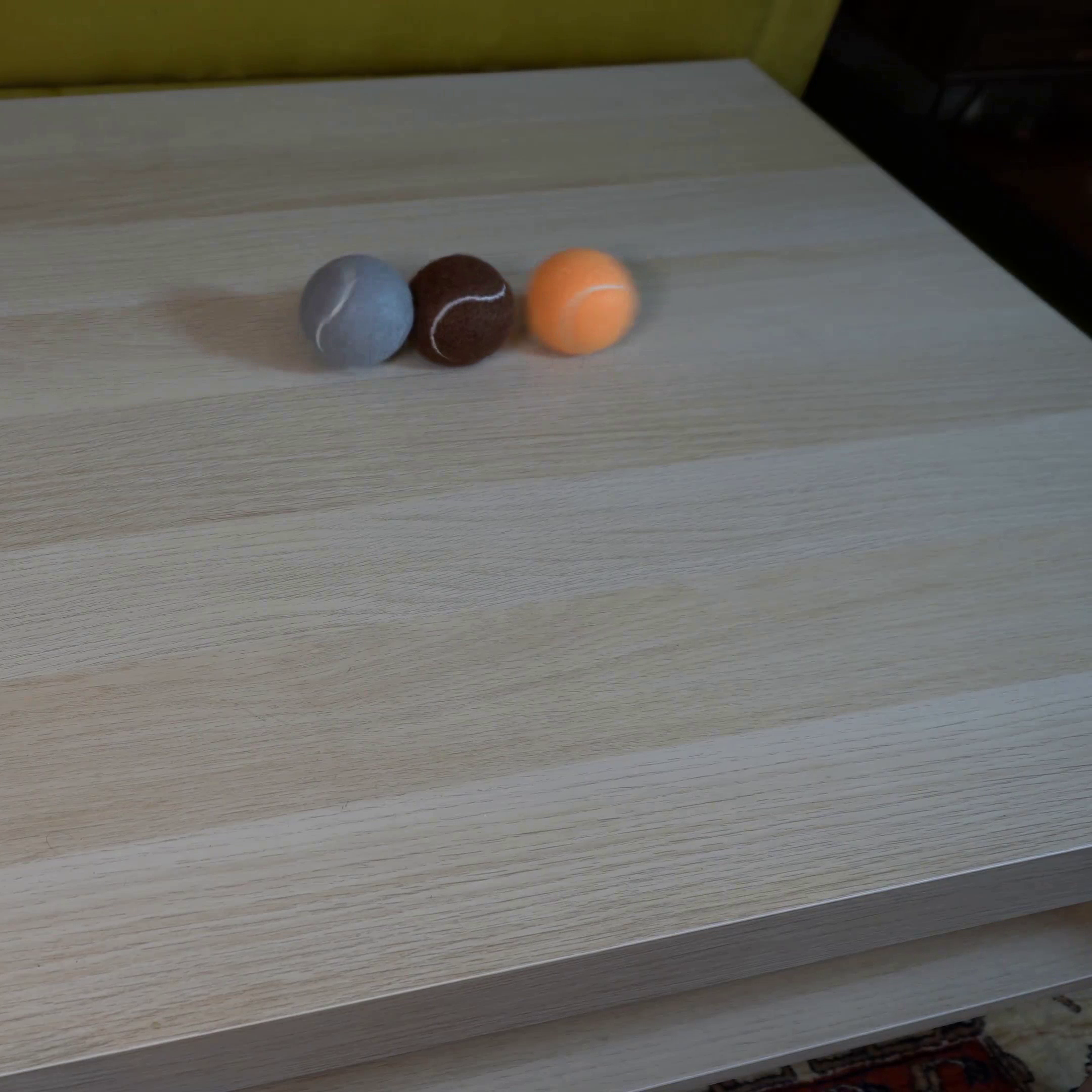}}&%
    \parbox[c]{\QualitativeCompareBoxWidth}{\includegraphics[width=\QualitativeCompareImageWidth]{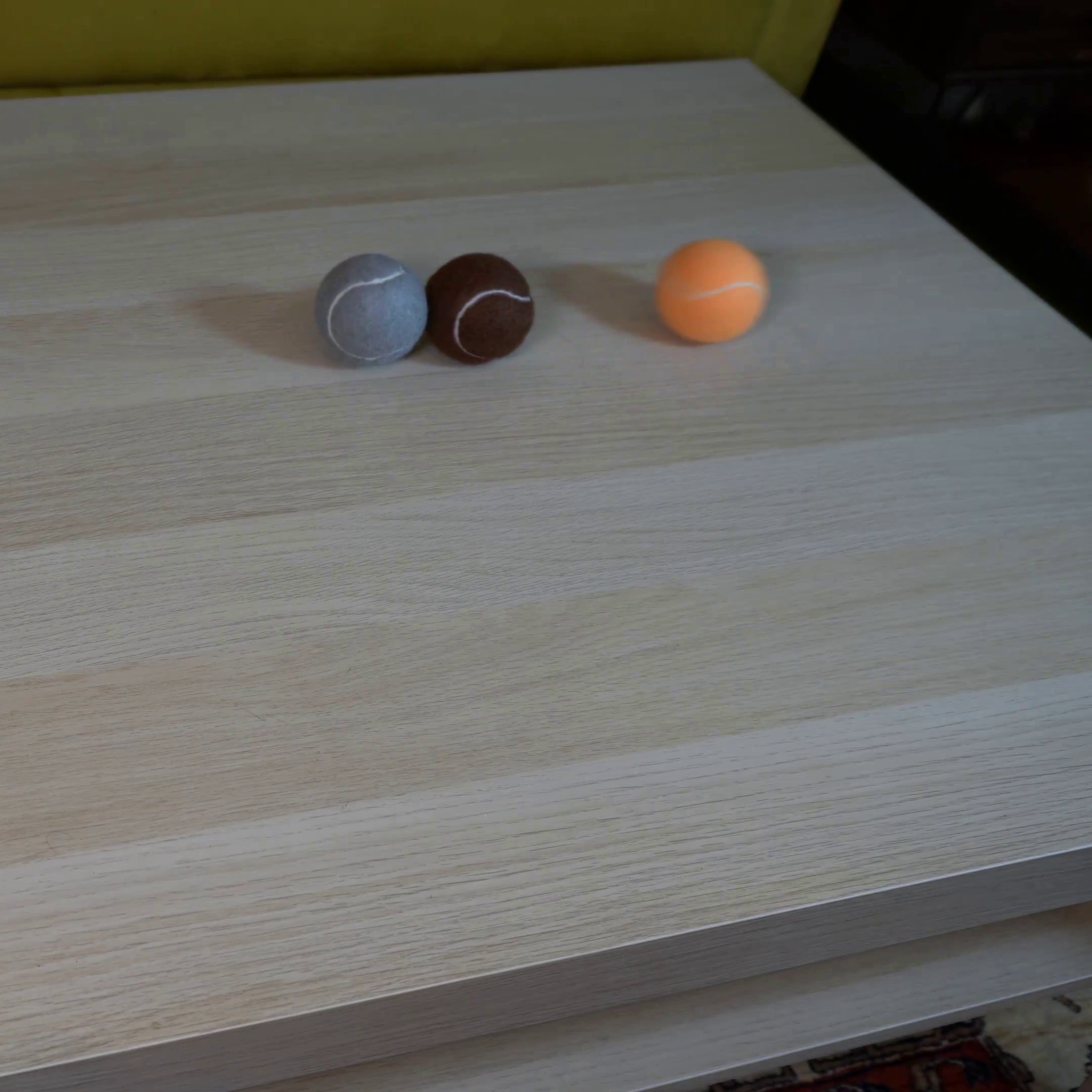}}&%
    \parbox[c]{\QualitativeCompareBoxWidth}{\includegraphics[width=\QualitativeCompareImageWidth]{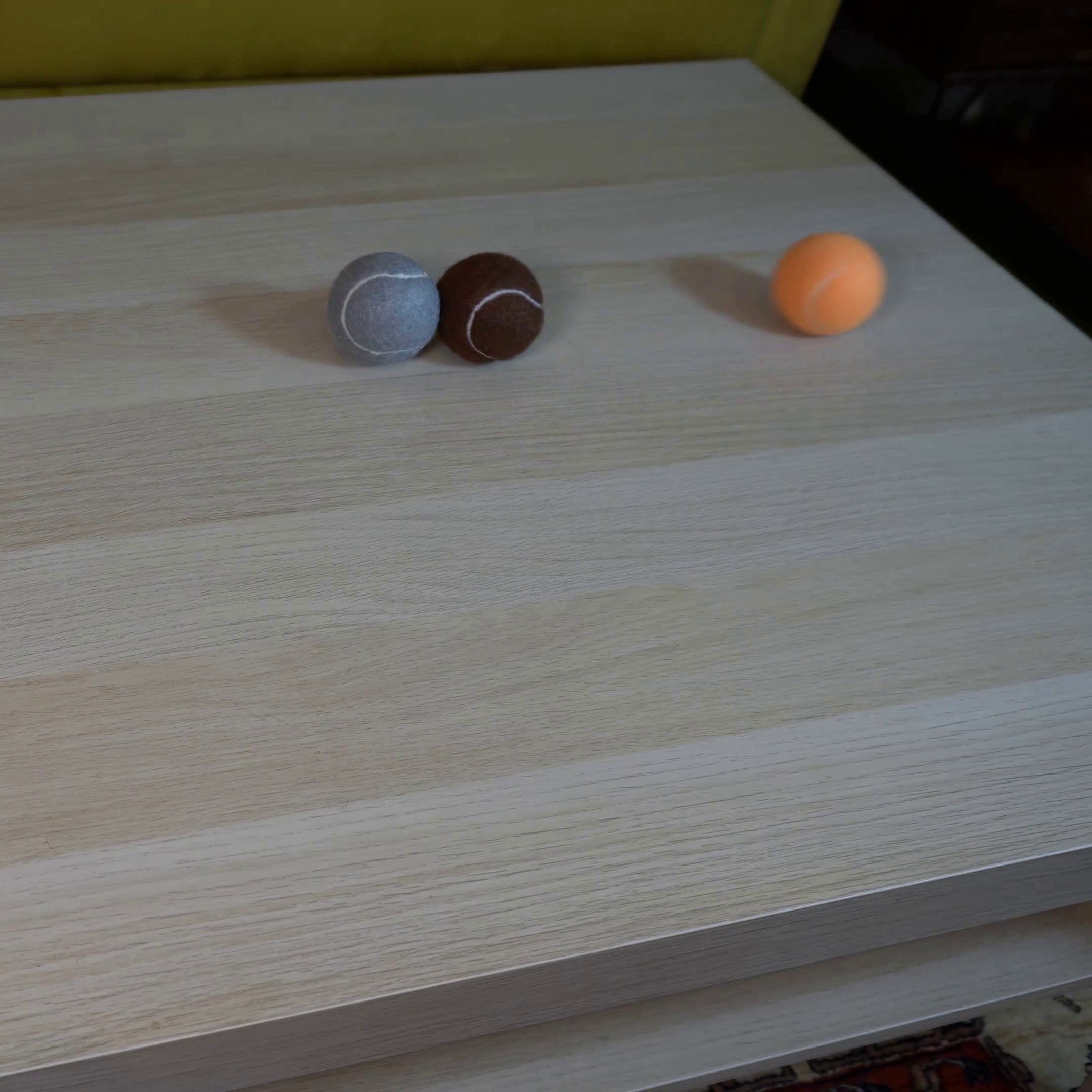}}%
    \\ \noalign{\vspace{2pt}} 

    \begin{tabular}{c} \rotatebox{90}{ \textsc{Human}} \end{tabular}&%
    \parbox[c]{\QualitativeCompareBoxWidth}{\includegraphics[width=\QualitativeCompareImageWidth]{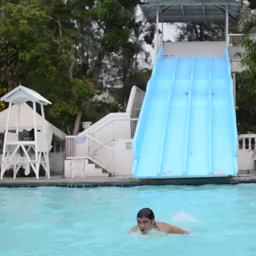}}&%
    \parbox[c]{\QualitativeCompareBoxWidth}{\includegraphics[width=\QualitativeCompareImageWidth]{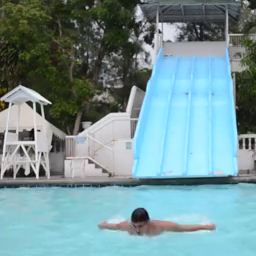}}&%
    \parbox[c]{\QualitativeCompareBoxWidth}{\includegraphics[width=\QualitativeCompareImageWidth]{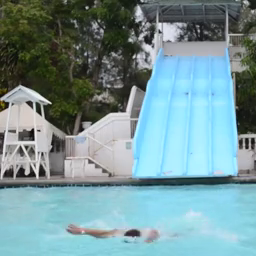}}&%
    \parbox[c]{\QualitativeCompareBoxWidth}{\includegraphics[width=\QualitativeCompareImageWidth]{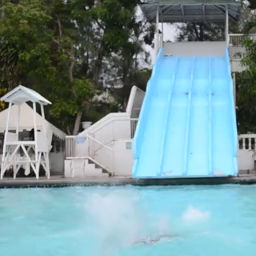}}&%
    \parbox[c]{\QualitativeCompareBoxWidth}{\includegraphics[width=\QualitativeCompareImageWidth]{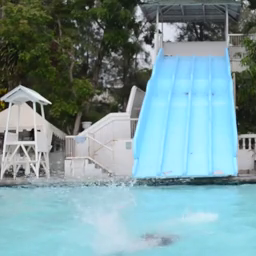}}%
    \\ \noalign{\vspace{2pt}} 

    \begin{tabular}{c} \rotatebox{90}{ \textsc{Animal}} \end{tabular}&%
    \parbox[c]{\QualitativeCompareBoxWidth}{\includegraphics[width=\QualitativeCompareImageWidth]{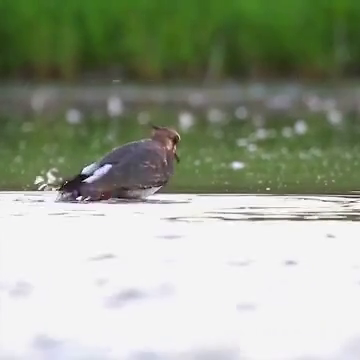}}&%
    \parbox[c]{\QualitativeCompareBoxWidth}{\includegraphics[width=\QualitativeCompareImageWidth]{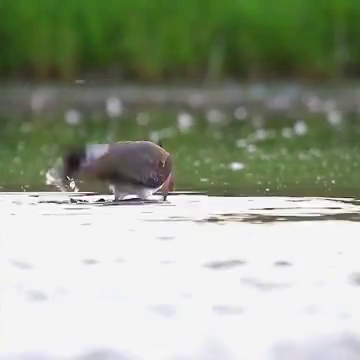}}&%
    \parbox[c]{\QualitativeCompareBoxWidth}{\includegraphics[width=\QualitativeCompareImageWidth]{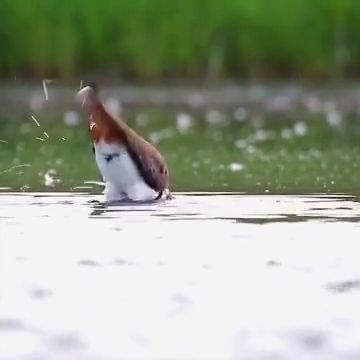}}&%
    \parbox[c]{\QualitativeCompareBoxWidth}{\includegraphics[width=\QualitativeCompareImageWidth]{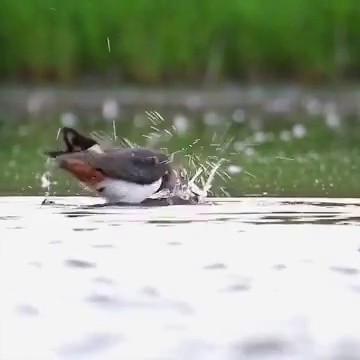}}&%
    \parbox[c]{\QualitativeCompareBoxWidth}{\includegraphics[width=\QualitativeCompareImageWidth]{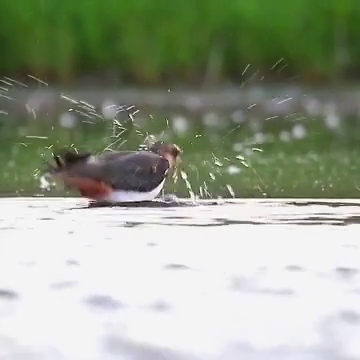}}%
    \\ 
    \noalign{\vspace{4pt}}
    & \multicolumn{5}{l}{%
        \begin{tikzpicture}
            \draw[-{Stealth[scale=1.2]}, line width=0.8pt] (0,0) -- (0.9\textwidth - 30pt, 0) 
                node[right, inner sep=2pt, font=\bfseries] {time};
        \end{tikzpicture}%
    } \\

    \end{tabular}%
    }
    \caption{
    \textbf{Qualitative results of our four diverse causal domains.} Each row illustrates the featured video in the subsets.
    }
    \label{fig:dataset_videos}
\vspace{-3mm}
\end{figure}

Our benchmark dataset $\mathcal{D}$ comprises four thematic subsets, each sourced from a well-established real-world video collection. We describe the construction protocol for each subset below, followed by a discussion of the dataset's overall scale and extensibility.

\noindent\textbf{General Subset $(\mathcal{D}_\text{General})$.}
We randomly sample 500 videos from the Moments in Time dataset~\cite{monfort2019moments}, retaining each clip at its original duration of 3 seconds. Moments in Time~\cite{monfort2019moments} is a large-scale collection of approximately one million 3-second clips curated by MIT, covering 339 action and event categories that span human activities, animal behaviors, natural phenomena, and object state changes. Its broad semantic diversity makes it well-suited for evaluating general everyday causality without domain-specific bias.

\noindent\textbf{Physics Subset $(\mathcal{D}_\text{Physics})$.}
We source 132 videos from the Physics IQ dataset~\cite{motamed2026generative}, selecting one camera viewpoint per scene from the dataset's 132 physical scenarios and trimming each clip to the first 5 seconds following the official evaluation protocol. Physics IQ is specifically designed to assess intuitive physical understanding in video generative models, covering phenomena such as gravity, fluid motion, and collision dynamics. Each clip presents an unambiguous physical event, making this subset particularly rich in temporally asymmetric causal cues.

\noindent\textbf{Human Action Subset $(\mathcal{D}_\text{Human})$.}
We randomly sample one video per action class from Kinetics-400~\cite{kay2017kinetics}, yielding 400 clips in total, each trimmed to the first 3 seconds. Kinetics-400 is a canonical human action recognition benchmark curated by DeepMind, comprising approximately 300K YouTube clips across 400 action categories such as swimming, playing guitar, and handshaking. The abundance of diverse human-centric activities provides strong coverage of goal-directed, agent-caused interactions where the causal arrow is tied to purposeful body motion.

\noindent\textbf{Animal Action Subset $(\mathcal{D}_\text{Animal})$.}
We randomly sample 200 videos from the Animal Kingdom dataset~\cite{ng2022animal}, trimming each clip to the first 3 seconds. Animal Kingdom is a large-scale dataset annotated with a rich taxonomy of animal behaviors, including foraging, aggression, courtship, and locomotion, across a highly diverse set of species. This subset extends our benchmark beyond anthropocentric scenarios to assess whether causal cognition generalizes to non-human agents and natural world dynamics.

\noindent\textbf{Scale and Extensibility.}
As discussed in the main paper, our benchmark is inherently extensible, as any real-world video can be incorporated at zero additional cost via temporal reversal. In the current work, we evaluate a total of 1,232 videos across the four subsets described above. This scale was chosen under practical computational constraints: evaluating 13 video diffusion models (several with billions of parameters) requires substantial GPU overhead, and the computational resource we can afford only supports us to evaluate with thousands of videos. Future researchers with access to greater resources, or those benchmarking a single target model, are encouraged to expand the dataset's scale and diversity by incorporating additional domains (\eg tool use, cooking, or sports) to keep the benchmark evolving alongside model capabilities.

A summary of the subset composition is provided in ~\cref{tab:dataset_setting}.

\begin{table}[h]
\centering
\caption{
\textbf{Composition of the YoCausal benchmark dataset.} Each subset is sourced from an existing real-world video collection, with clips uniformly trimmed to the specified duration.}
\label{tab:dataset_setting}
\begin{tabular}{lcccc}
\toprule
\textbf{Subset} & \textbf{Source Dataset} & \textbf{Clip Duration} & \textbf{\# Videos} \\
\midrule
$\mathcal{D}_\text{General}$ & Moments in Time~\cite{monfort2019moments} & 3 s & 500 \\
$\mathcal{D}_\text{Physics}$ & Physics IQ~\cite{motamed2026generative}   & 5 s (first 5 s) & 132 \\
$\mathcal{D}_\text{Human}$   & Kinetics-400~\cite{kay2017kinetics}       & 3 s (first 3 s) & 400 \\
$\mathcal{D}_\text{Animal}$  & Animal Kingdom~\cite{ng2022animal}        & 3 s (first 3 s) & 200 \\
\midrule
\textbf{Total}  & ---                                       & ---             & \textbf{1,232} \\
\bottomrule
\end{tabular}
\end{table}
\subsection{Models Setting}
\label{suppl_sec:model_setting}

We evaluate 13 state-of-the-art open-source text-to-video diffusion models spanning diverse architectures and parameter scales: AnimateDiff-SDv1.5/SDXL~\cite{guo2023animatediff}, CogVideoX-2B/5B~\cite{hong2022cogvideo, yang2024cogvideox}, CogVideoX1.5-5B~\cite{hong2022cogvideo, yang2024cogvideox}, Mochi-1-preview~\cite{genmo2024mochi}, HunyuanVideo~\cite{kong2024hunyuanvideo}, Wan2.1-T2V-1.3B/14B~\cite{wan2025wan}, Wan2.2-TI2V-5B/T2V-A14B~\cite{wan2025wan}, and LTX-Video-2B/13B~\cite{hacohen2024ltx}.

To ensure each model operates under its optimal conditions, all inference configurations strictly follow the official recommended defaults for each model (\cref{tab:model_spec}), including output resolution, number of frames, and frames per second (FPS). It is worth noting that classifier-free guidance (CFG)~\cite{ho2022classifier} is not applied during evaluation: rather than performing full denoising generation, we directly compute the MSE between the model's predicted noise $\hat{\epsilon}$ and the sampled Gaussian noise $\epsilon$ at each timestep. In addition, we report model specifications (\eg spatial compression ratios) in \cref{tab:model_spec} to further investigate the relationship between model features and causal cognition performance.

\begin{table}[ht]
    \centering
    \caption{
    \textbf{Specifications of evaluated video generation models, grouped by model series.}
    For each model, we report the parameter count, prediction target, spatial and temporal compression ratios of the VAE, frame window size, frames per second (FPS), and default output resolution.
    }
    \label{tab:model_spec}
    \resizebox{\textwidth}{!}{%
    \begin{tabular}{l c c c c c c c}
        \toprule
        Model Name & Params. & \makecell{Prediction \\ Target} & \makecell{Spatial \\ Compress} & \makecell{Temporal \\ Compress} & \makecell{Frame \\ Window} & FPS & \makecell{Default \\ Resolution} \\
        \midrule
        \multicolumn{8}{l}{\textbf{Wan}} \\
        \enskip Wan 2.2 A14B & 14B & Flow & $8\times$ & $4\times$ & 81 & 16 & $1280 \times 720$ \\
        \enskip Wan 2.2 5B & 5B & Flow & $8\times$ & $4\times$ & 121 & 24 & $1280 \times 720$ \\
        \enskip Wan 2.1 14B & 14B & Flow & $8\times$ & $4\times$ & 81 & 16 & $1280 \times 720$ \\
        \enskip Wan 2.1 1.3B & 1.3B & Flow & $8\times$ & $4\times$ & 81 & 16 & $832 \times 480$ \\
        \midrule
        \multicolumn{8}{l}{\textbf{Hunyuan}} \\
        \enskip HunyuanVideo & 13B & Flow & $8\times$ & $4\times$ & 129 & 24 & $1280 \times 720$ \\
        \midrule
        \multicolumn{8}{l}{\textbf{Mochi}} \\
        \enskip Mochi 1 & 10B & Flow & $8\times$ & $6\times$ & 163 & 30 & $848 \times 480$ \\
        \midrule
        \multicolumn{8}{l}{\textbf{CogVideoX}} \\
        \enskip CogVideoX-5B & 5B & $v$ & $8\times$ & $4\times$ & 81 & 16 & $720 \times 480$ \\
        \enskip CogVideoX-2B & 2B & $v$ & $8\times$ & $4\times$ & 81 & 16 & $720 \times 480$ \\
        \enskip CogVideoX1.5-5B & 5B & $v$ & $8\times$ & $4\times$ & 81 & 16 & $1360 \times 768$ \\
        \midrule
        \multicolumn{8}{l}{\textbf{AnimateDiff}} \\
        \enskip AnimateDiff SD 1.5 & 1.4B & $\epsilon$ & $8\times$ & $1\times$ & 16 & 8 & $512 \times 512$ \\
        \enskip AnimateDiff SDXL & 3.5B & $\epsilon$ & $8\times$ & $1\times$ & 16 & 8 & $1024 \times 1024$ \\
        \midrule
        \multicolumn{8}{l}{\textbf{LTXV}} \\
        \enskip LTXV-13B-0.9.8 & 13B & Flow & $32\times$ & $8\times$ & 121 & 30 & $1216 \times 704$ \\
        \enskip LTXV-2B-0.9.6 & 2B & Flow & $32\times$ & $8\times$ & 121 & 30 & $1216 \times 704$ \\
        \bottomrule
    \end{tabular}}
\end{table}

\subsection{Video Preprocessing}
\label{suppl_sec:video_preprocessing}

Since each model operates under its officially recommended settings (~\cref{suppl_sec:model_setting}), input specifications inevitably differ across models in spatial resolution, frame rate, and temporal length. To ensure a fair and consistent evaluation, we apply a unified preprocessing pipeline encompassing three stages: resolution adaptation, FPS resampling, and long video handling.

\noindent\textbf{Resolution Adaptation.}
Since different models require different input resolutions and aspect ratios, we employ two adaptation strategies depending on the model's design. For models that only support a fixed resolution, we first rescale the input video such that its shorter side matches the target dimension, and then apply a center crop to obtain the model's officially specified resolution and aspect ratio. For models that support multiple aspect-ratio buckets (\eg the Wan series and HunyuanVideo), we automatically select the bucket resolution closest to the original video's aspect ratio.

\noindent\textbf{FPS Resampling.}
We use FFmpeg to resample all source videos to each model's prescribed frame rate, ensuring that the temporal sampling rate is consistent with the model's training configuration.

\noindent\textbf{Long Video Handling.}
Different models have different maximum frame windows (\ie the number of frames they can process in a single forward pass). When the total number of frames in a source video exceeds a model's frame window limit, naively truncating the video would compromise evaluation fairness, as the model would only observe a partial segment. To address this, we partition the video along the temporal axis into consecutive frame-window-sized segments. If the last segment is shorter than a full window, we pad it by prepending the necessary number of frames from the preceding segment as temporal context. Importantly, this context portion is used solely to provide temporal information and is excluded from the denoising loss accumulation. Finally, we sum the denoising losses across all windows for the entire video, and use this aggregated value as the basis for comparing forward and reversed sequences.

\subsection{Details on RSI Algorithm}
\label{suppl_sec:RSI}

The Reverse Surprise Index (RSI) quantifies a model's ability to perceive the arrow of time, serving as our Level-1 metric. The complete procedure is formalized in~\cref{alg:RSI} (Stage~1 and Stage~2).

\noindent\textbf{Timestep and Noise Sampling.}
The denoising loss is defined as an expectation over both diffusion timesteps $t$ and sampled noise $\epsilon$. To approximate this expectation, we uniformly sample $K{=}10$ timesteps from $[1, T]$, excluding the boundaries $t{=}0$ and $t{=}T_{\max}$ (fully clean and fully noised states). For each timestep, we draw $N_{\epsilon}{=}1$ noise sample $\epsilon \sim \mathcal{N}(\mathbf{0}, \mathbf{I})$ and apply it identically to both the forward sequence $x^f$ and the reversed sequence $x^r$, ensuring that the two versions face identical denoising difficulty. For latent diffusion models, the noise is sampled in the latent space; for pixel-space models, it is sampled at the full spatial resolution.

Following LikePhys\cite{yuan2025likephys}, we condition both the forward and reversed sequences on the same text prompt, \ie the original caption of the forward video in the dataset.
Writing a separate caption for the reversed clip would conflate causal understanding with the ability to follow unrealistic instructions. While a null prompt is also viable, conditioning on the meaningful caption provides a sharper denoising signal at low-SNR timesteps, avoiding the randomness of this evaluation method and giving more stable evaluation process. We also empirically validate that RSI and CCI remain significant under null prompts (see \cref{suppl_sec:prompt_bias}), confirming that our metrics capture genuine temporal-causal structure rather than text-video misalignment artifacts.

We set $K{=}10$ and $N_{\epsilon}{=}1$ primarily due to computational constraints: our evaluation spans 13 large-scale models over 1{,}232 videos, making exhaustive sampling prohibitive given our limited computational resources. As noted in~\cref{suppl_sec:dataset}, future researchers with sufficient computational resources are encouraged to increase both $K$ and $N_{\epsilon}$, yielding estimates closer to the true expected denoising loss.

\noindent\textbf{Decision Criterion.}
For each video, we average the denoising losses over all $K$ timesteps and $N_{\epsilon}{=}1$ noise samples. If $\mathcal{L}(\theta;\, x^f) < \mathcal{L}(\theta;\, x^r)$, the model is deemed to have correctly identified the temporal direction. RSI is then the proportion of videos satisfying this condition; $50\%$ corresponds to chance level.

\begin{algorithm}[h!]
\small                  
\setlength{\itemsep}{0pt}
\caption{YoCausal: Two-Level Causality Evaluation for Video Diffusion Models}
\label{alg:RSI}
\begin{algorithmic}[1]

\REQUIRE Video dataset $\mathcal{D} = \{\mathcal{D}_1, \ldots, \mathcal{D}_n\}$;
         pretrained VDM $\epsilon_\theta$;
         $K$ uniformly sampled timesteps;
         $N_\epsilon$ noise samples per timestep;
         VLM causality classifier $\mathcal{C}_\text{vlm}$
\ENSURE  $\text{RSI}(\mathcal{D})$, $\text{CCI}(\mathcal{D})$

\STATE \textbf{// Stage 1: Compute Denoising Losses}
\FOR{each $\mathcal{D}_i \in \mathcal{D}$, each video $x \in \mathcal{D}_i$}
  \STATE Set $x^f \leftarrow x$;\; $x^r \leftarrow \textsc{Temporal-Reverse}(x)$
  \FOR{each direction $\tilde{x} \in \{x^f, x^r\}$}
    \STATE $\mathcal{L}(\tilde{x}) \leftarrow 0$;\; sample $\{t_k\}_{k=1}^{K}$ uniformly from $[1,T]$
    \FOR{$k = 1$ \TO $K$, $m=1$ \TO $N_{\epsilon}$}
      \STATE Sample $\epsilon \sim \mathcal{N}(\mathbf{0}, \mathbf{I})$
      \STATE $\tilde{x}_{t_k} \leftarrow \sqrt{\bar{\alpha}_{t_k}}\,\tilde{x} + \sqrt{1-\bar{\alpha}_{t_k}}\,\epsilon$
      \STATE $\hat{\epsilon} \leftarrow \epsilon_\theta(\tilde{x}_{t_k}, t_k)$;\;
             $\mathcal{L}(\tilde{x}) \mathrel{+}= \|\epsilon - \hat{\epsilon}\|_2^2$
    \ENDFOR
    \STATE $\mathcal{L}(\tilde{x}) \leftarrow \mathcal{L}(\tilde{x})\,/\, (K \times N_{\epsilon})$
  \ENDFOR
\ENDFOR

\STATE \textbf{// Stage 2: Level-1 — Reverse Surprise Index (RSI)}
\FOR{each $\mathcal{D}_i \in \mathcal{D}$}
  \STATE $S_i \leftarrow \sum_{x \in \mathcal{D}_i} \mathbf{1}\!\left[\mathcal{L}(x^r) > \mathcal{L}(x^f)\right]$;\;
         $\text{RSI}(\mathcal{D}_i) \leftarrow S_i\,/\,|\mathcal{D}_i|$
\ENDFOR
\STATE $\text{RSI}(\mathcal{D}) \leftarrow \frac{1}{|\mathcal{D}|}\sum_{\mathcal{D}_i \in \mathcal{D}} \text{RSI}(\mathcal{D}_i)$

\STATE \textbf{// Stage 3: VLM Causality Stratification}
\STATE $\mathcal{D}_c \leftarrow \emptyset$;\; $\mathcal{D}_{nc} \leftarrow \emptyset$
\FOR{each $\mathcal{D}_i \in \mathcal{D}$, each $x \in \mathcal{D}_i$}
  \IF{$\mathcal{C}_\text{vlm}(x) = \textsc{True}$}
    \STATE $\mathcal{D}_c \leftarrow \mathcal{D}_c \cup \{x\}$
  \ELSE
    \STATE $\mathcal{D}_{nc} \leftarrow \mathcal{D}_{nc} \cup \{x\}$
  \ENDIF
\ENDFOR

\STATE \textbf{// Stage 4: Level-2 — Causality Cognition Index (CCI)}
\STATE Compute $\text{RSI}(\mathcal{D}_c)$ and $\text{RSI}(\mathcal{D}_{nc})$ as in Stage 2
\STATE $\text{CCI}(\mathcal{D}) \leftarrow \text{RSI}(\mathcal{D}_c) - \text{RSI}(\mathcal{D}_{nc})$

\RETURN $\text{RSI}(\mathcal{D})$,\; $\text{CCI}(\mathcal{D})$

\end{algorithmic}
\end{algorithm}

\subsection{Details on CCI Algorithm}
\label{suppl_sec:CCI}
\raggedbottom

As discussed in the main paper, in order to isolate the pure causal cognition signal from the overall arrow-of-time perception captured by RSI, we propose the Causality Cognition Index (CCI). It is defined as the difference in RSI between the causal and non-causal subsets of the dataset: $\text{CCI}(\mathcal{D}) = \text{RSI}(\mathcal{D}_c) - \text{RSI}(\mathcal{D}_{nc})$.
where $\mathcal{D}_c$ denotes the subset of videos containing salient cause-and-effect interactions, and $\mathcal{D}_{nc}$ denotes the subset without such interactions. The intuition behind this differential design is as follows: the surprise induced by reversing a non-causal video originates primarily from the reversal of the statistical arrow of time, whereas the surprise from reversing a causal video additionally includes the physical implausibility of inverted causality. By taking the difference, CCI cancels out the shared confounding factor: arrow-of-time sensitivity and isolates the model's independent sensitivity to causal violations. The complete procedure is formalized in ~\cref{alg:RSI} (Stage~4).

\begin{figure*}[h!]
    \centering
    \includegraphics[width=1.0\linewidth]{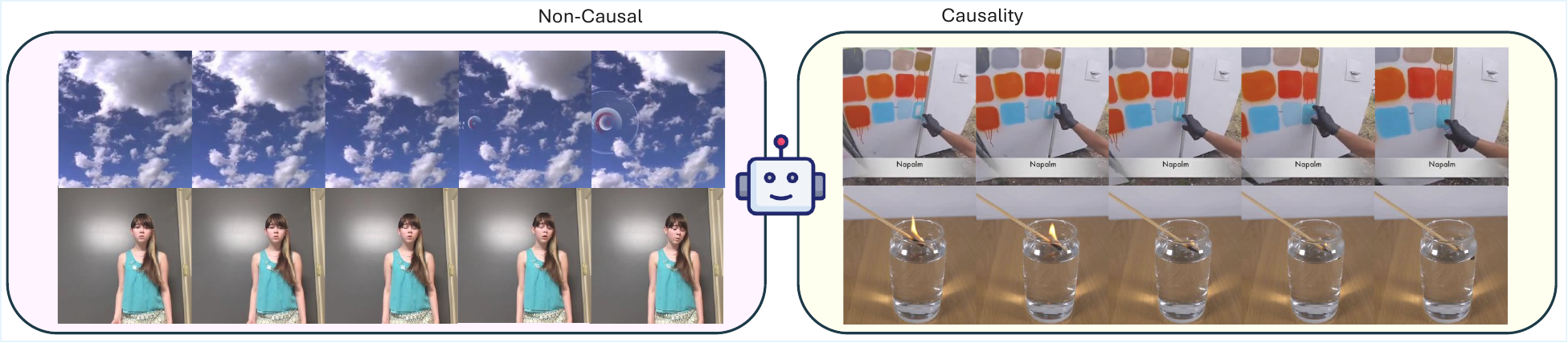}
    \caption{
    \textbf{Examples from the causal and non-causal subsets.}
    Non-causal videos (left) and Causal videos (right) are split by VLMs.
    }
    \label{fig:vlm_judge_video}
\end{figure*}

\noindent\textbf{VLM-Based Causal/Non-Causal Partitioning.}
Computing CCI requires partitioning the dataset $\mathcal{D}$ into $\mathcal{D}_c$ and $\mathcal{D}_{nc}$. To ensure the scalability of our benchmark, we employ an advanced Vision-Language Model (VLM) to automatically determine whether each video contains obvious causal interactions, rather than relying on manual annotation (~\cref{alg:RSI} (Stage~3)). The visualization result is shown in \cref{fig:vlm_judge_video}. The rationale for this automated strategy is that judging \emph{whether causality exists} in a video is a substantially easier task than judging \emph{whether a causal relation is correct}, making VLMs well-suited for this binary classification. In the main paper, we validate the reliability of this approach from three complementary perspectives, confirming that VLM-based labeling serves as a reliable proxy for human judgment.

\noindent\textbf{VLM Configuration and Prompt.}
We use Gemini 3.0 Pro~\cite{team2023gemini} as the VLM judge. Each video is provided as visual input alongside the following prompt:

\begin{tcolorbox}[title=Prompt, colback=gray!5, colframe=gray!50]
\scriptsize
\texttt{You are an expert in spatio-temporal video analysis. Your task is to analyze the provided video and determine whether it exhibits ``Causality.''}

\textbf{\# Definition of Causality}
In this context, ``Causality'' is defined strictly as observable cause-and-effect mechanisms within spatio-temporal dynamics.
\begin{itemize}
    \item It means Event A explicitly and visibly causes Event B.
    \item It goes beyond strict Newtonian physics (e.g., collisions or gravity) to include complex logical event sequences (e.g., a pencil moving on paper causing text to appear, or a person flipping a switch causing a light to turn on).
\end{itemize}

\textbf{\# Output Format}
Provide your final answer in the following JSON format:
\scriptsize
\begin{lstlisting}[basicstyle=\scriptsize\ttfamily, frame=none]
{
  "reasoning": "Step-by-step reasoning evaluating if this
  specific resulting_event is strictly dependent on the initiating_event.",
  "video_has_general_causality": true/false,
  "confidence": 1-5
}
\end{lstlisting}
\end{tcolorbox}

\subsection{Prompt Bias}
\label{suppl_sec:prompt_bias}

In our experimental design, both the forward and reversed videos are conditioned on the same text prompt, namely the original caption describing the forward video, avoiding the randomness under low-SNR timesteps with null prompt. A natural concern is whether the elevated denoising loss on reversed videos merely reflects a semantic mismatch between the visual content and the prompt, rather than the model's sensitivity to causal violations. We address this concern from an empirical ablation under a null-prompt setting.

\noindent\textbf{Empirical validation: null-prompt ablation.}
We design an ablation to empirically validate that our metrics are not driven by text-video misalignment.

\noindent\textit{Hypothesis.} If the elevated denoising loss on reversed videos primarily reflected text-video misalignment, then removing the prompt conditioning should eliminate the bulk of the RSI/CCI signal. Conversely, if the signal genuinely originates from the model's internalized temporal and causal cognition, then RSI and CCI under null prompts should be close to values under forward video prompt and preserve their discriminative power.

\noindent\textit{Setup.} We re-evaluate three representative models under a \emph{null-prompt} setting, where both forward and reversed sequences are conditioned on an empty caption.

\begin{table*}[h]
\small
\centering
\caption{\textbf{Null-prompt ablation results.} 
We compare RSI and CCI under the forward prompt setting (used in our main experiments) and the null prompt setting on three representative models from a range of model families and capabilities. Both metrics under null prompts are still close to the values under forward prompt, and the discriminative structure is preserved, supporting our claim that RSI and CCI are not artifacts of text-video misalignment.}
\setlength{\tabcolsep}{8pt}
\resizebox{0.95\textwidth}{!}{%
  \begin{tabular}{@{}lccc@{}}
  \toprule
   & HunyuanVideo
   & CogVideoX-5b
   & LTX-Video-2b-0.9.6 \\
  \midrule
  RSI w/ forward prompt
   & 52.05\% & 49.92\% & 58.86\% \\
  RSI w/ null prompt
   & 51.17\% & 47.55\% & 56.92\% \\
  \midrule
  CCI w/ forward prompt
   & -0.29\% & 5.09\% & -0.20\% \\
  CCI w/ null prompt
   & -2.95\% & 6.17\% & -0.30\% \\
  \bottomrule
  \end{tabular}%
}%
\label{tab:null_prompt}
\end{table*}

\noindent\textit{Results.} As reported in \cref{tab:null_prompt}, removing the prompt produces only negligible changes in RSI and CCI. Crucially, the discriminative structure of both metrics is preserved, meaning the sign remain consistent with the forward-prompt setting.

This invariance directly falsifies the misalignment hypothesis. 
If text-video misalignment were the dominant source 
of elevated reversed-video loss, removing the prompt would pull all index to randomness line across all models. The observed pattern is incompatible.
The dominant signal originates from the 
model's internalized cognition, 
not from prompt conditioning.
This empirically complements our argument: the forward prompt 
serves as a stable, easily reproducible conditioning choice for the main benchmark, and is not the source of the cognition signal we measure.

We think using our benchmark with null prompt is also a reasonable choice, it may cancel out the slight text-misalignment but introduces randomness at low-SNR timesteps. In contrast, using our benchmark with forward prompt will cancel out the randomness at low-SNR timesteps but introduces little text-misalignment. That is a trade-off and in this paper we choose the more stable one (with forward prompt).

\subsection{Details on Human Annotating}
\label{suppl_sec:human_annotating}

To establish a human upper bound for the YoCausal benchmark, we recruit human annotators to label the temporal direction of all 1{,}232 videos. The labeling protocol is designed to maintain methodological parity with the Reverse Surprise Index (RSI) evaluation pipeline applied to models. Below, we describe the annotation procedure and the design of the ``Unknown'' option.

\noindent\textbf{Annotation Procedure and Decision Protocol.}
As shown in \cref{fig:human_labeling_web}, annotators first read the corresponding text prompt, ensuring that human judgments are made under the same semantic constraints as model evaluations. Each annotator then sequentially watches the forward and reverse versions of the same video, presented in randomized order, and after viewing both clips, determines which one is the reversed version.

The rationale behind this sequential design mirrors how VDMs process the two versions: rather than simultaneously comparing both videos side by side, a VDM independently computes the denoising loss for each direction and infers the temporal order by comparing the resulting loss values, which is indirect comparison through its internal learned prior. Our annotation protocol follows the same principle: annotators watch each version independently at different pages and form an internal judgment about the degree of abnormality in each clip. The final decision of which version is reversed is then based on comparing these two independently perceptions of anomaly. The human annotator's internal sense of abnormality functionally corresponds to the model's denoising loss.

Furthermore, to direct annotators' attention toward high-level causal reasoning rather than low-level visual artifacts, we limit viewing to at most three replays for each of the forward and reverse versions.

\noindent\textbf{Design and Modeling of the ``Unknown'' Option.}
In the real world, certain events exhibit negligible directional cues (\eg static scenes, periodic motions), making it inherently difficult to distinguish forward from reversed playback. To deal with such cases, the annotation interface provides an ``Unknown'' option, preventing annotators from being forced into arbitrary decisions. Across all 1{,}232 samples, approximately 20\% of videos are marked as ``Unknown.''

To maintain consistency with model behavior under uncertainty, samples labeled as ``Unknown'' are assigned an expected win rate of 0.5 (equivalent to the random-guess baseline) when computing the overall RSI score.

We provide a screenshot of the annotation web interface in ~\cref{fig:human_labeling_web}:

\begin{figure}[h!]
    \centering
    \small
    \resizebox{\textwidth}{!}{
    \begin{tabular}{@{}c@{}c@{}c@{}}
    \includegraphics[height=4cm]{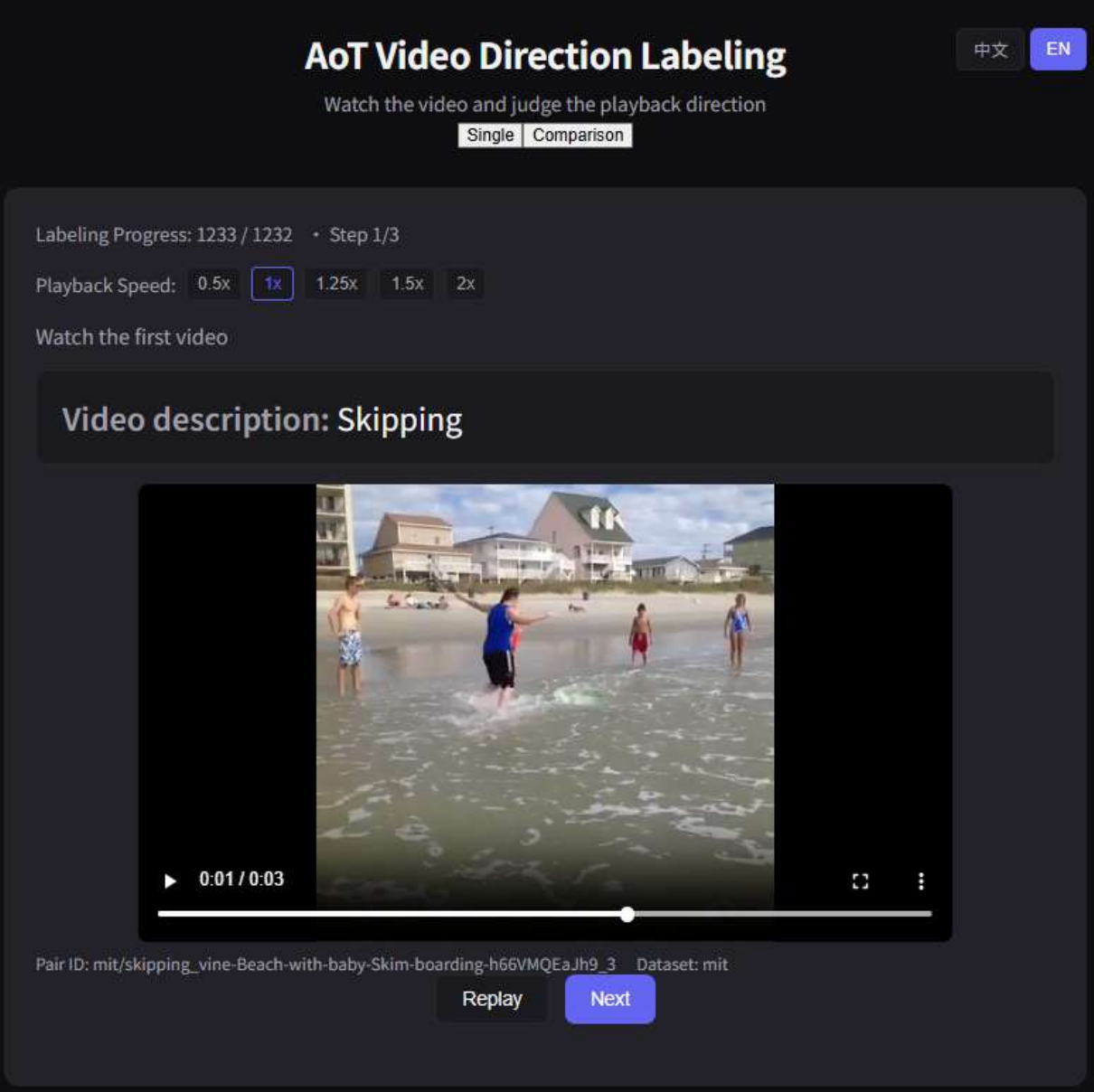} & 
    \includegraphics[height=4cm]{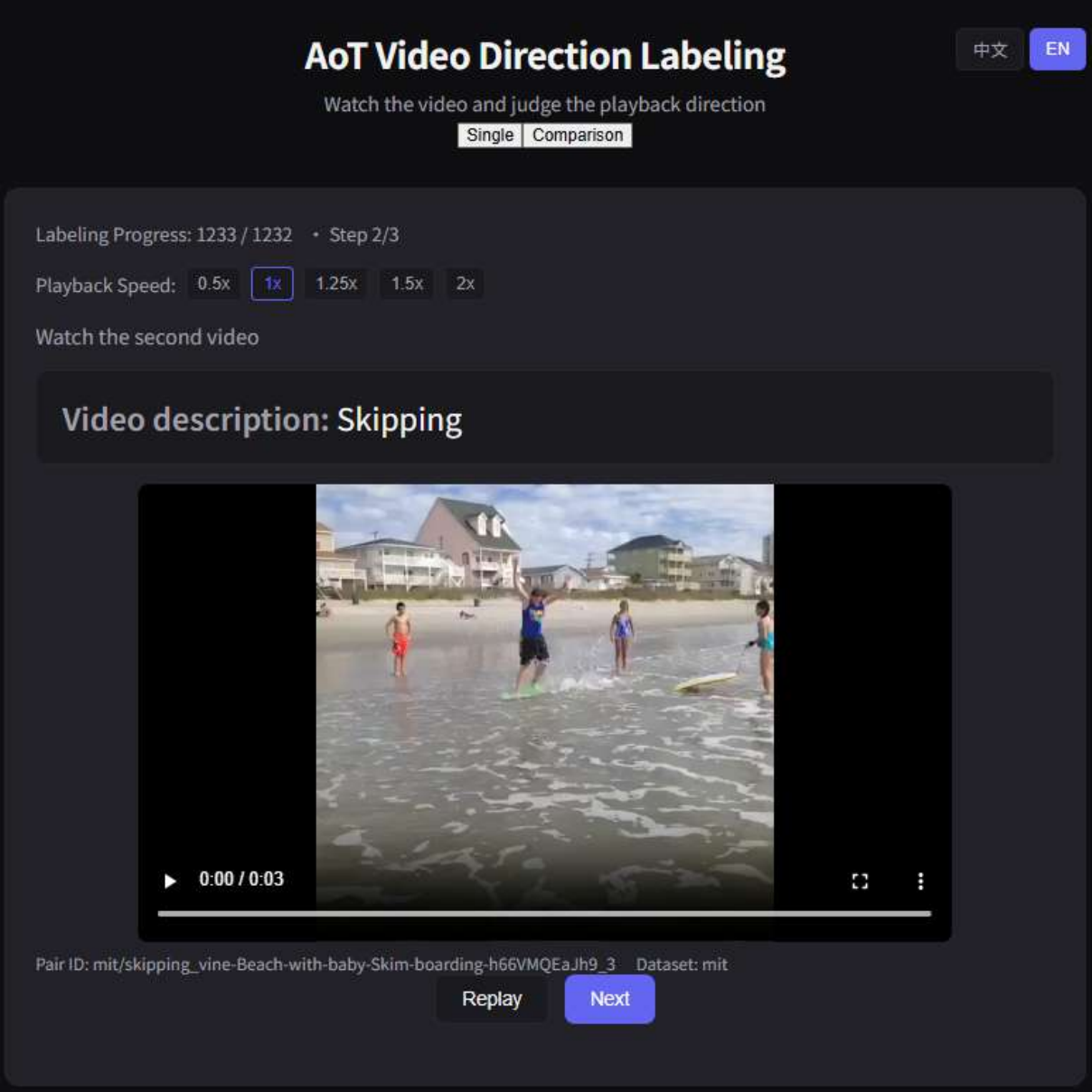} & 
    \includegraphics[height=4cm]{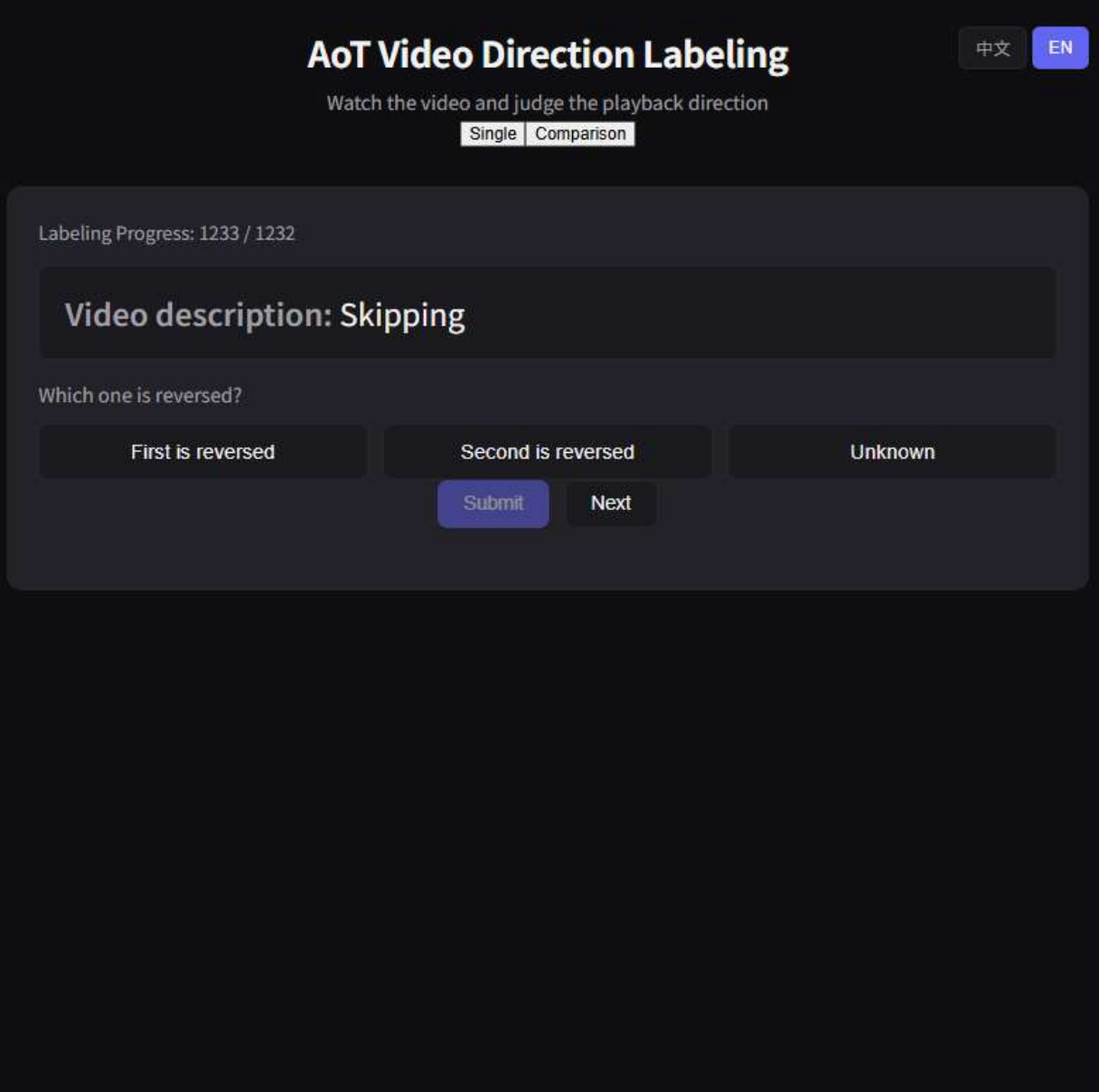}
    \end{tabular}
    }
    \caption{
    \textbf{Website for human annotation.}
    Annotators are shown a text prompt (top) and sequentially watch two versions of the same video—forward and reversed—presented in randomized order (left and middle panels). After viewing both clips, annotators select which version is reversed, with an ``Unknown'' option available.
    }
    \label{fig:human_labeling_web}
\end{figure}
\subsection{Details on Human Preference}
\label{suppl_sec:human_preference}

To verify the consistency between YoCausal's evaluation results and human subjective judgments, we design a \emph{Human Causality Preference} user study. The study encompasses model selection, prompt curation, evaluation procedure, and scoring, each of which is detailed below.

\noindent\textbf{Model Selection.}
Since the study adopts a pairwise comparison protocol, participants only need to compare two videos at a time. This allows us to include all 13 evaluated models rather than a subset of representative models, so that the human ranking can be directly compared with the full YoCausal ranking.

\noindent\textbf{Prompt Curation.}
We select 15 prompts with relatively identifiable causal interactions from each of the four subsets ($\mathcal{D}_{\mathit{General}}$, $\mathcal{D}_{\mathit{Physics}}$, $\mathcal{D}_{\mathit{Human}}$, $\mathcal{D}_{\mathit{Animal}}$), yielding a total of 60 prompts. Each of the 13 models then generates a video for every prompt, producing $13 \times 60 = 780$ videos in total.

\noindent\textbf{Evaluation Procedure.}
We deploy an online user-study website for the evaluation. On each page, the website presents two videos generated by two different models from the same prompt, displayed side by side with randomized left/right placement. Participants are asked to judge which video depicts more plausible causal interactions on a 5-point scale: \emph{A clearly better}, \emph{A slightly better}, \emph{About the same}, \emph{B slightly better}, and \emph{B clearly better}. Since videos generated by different models may exhibit similar levels of causal plausibility, the \emph{About the same} option avoids forcing artificial distinctions. We recruit 40 participants in total, and each participant completes 25 comparisons, yielding 1,000 pairwise judgments. The comparisons are assigned so that all $\binom{13}{2}=78$ model pairs are covered, each pair is compared on multiple distinct prompts.

We provide a screenshot of the human preference web interface in \cref{fig:user_study_web}:
\begin{center}
    \includegraphics[width=0.9\textwidth]{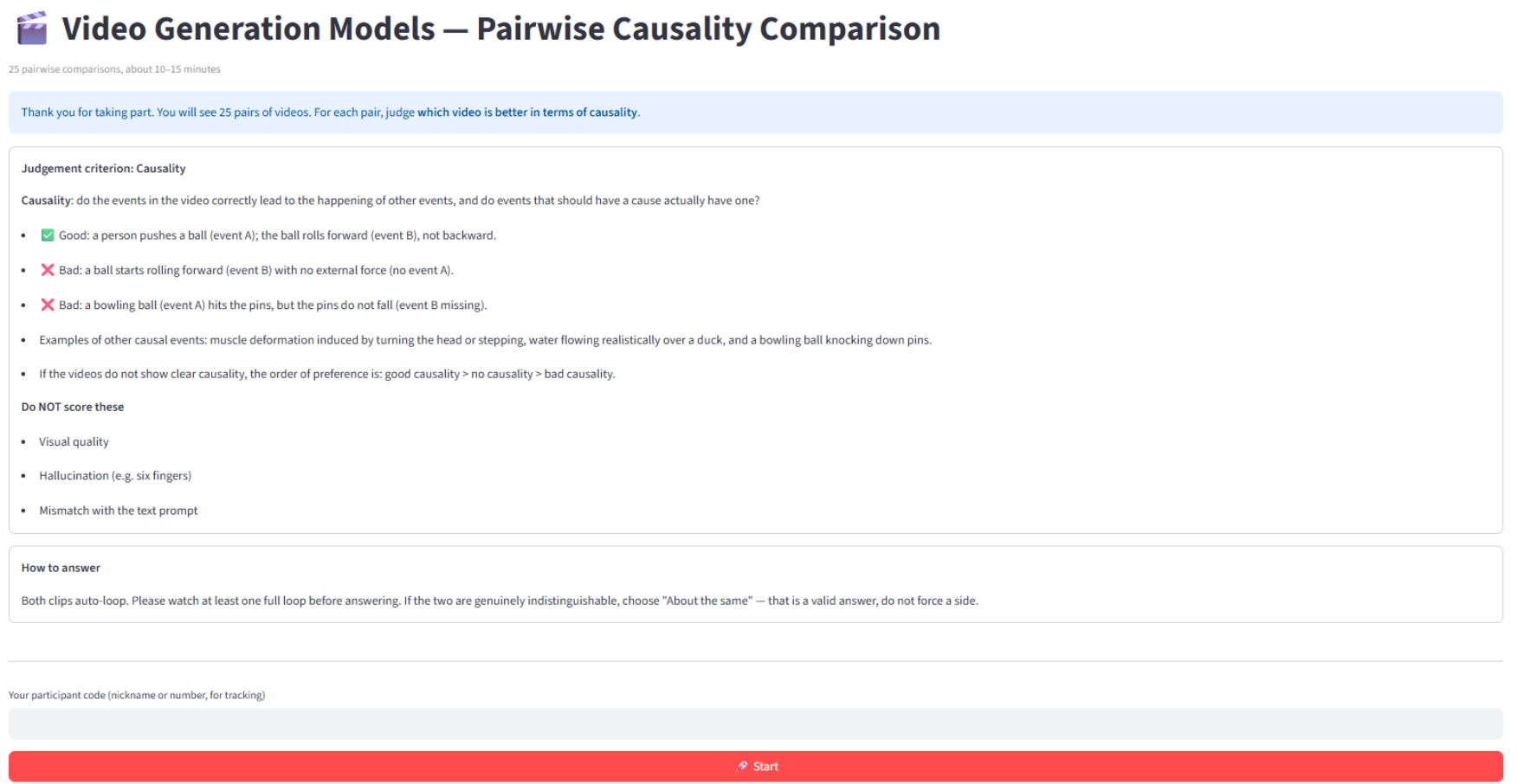}
    \captionof*{figure}{(a)}
\end{center}

\begin{center}
    \includegraphics[width=0.9\textwidth]{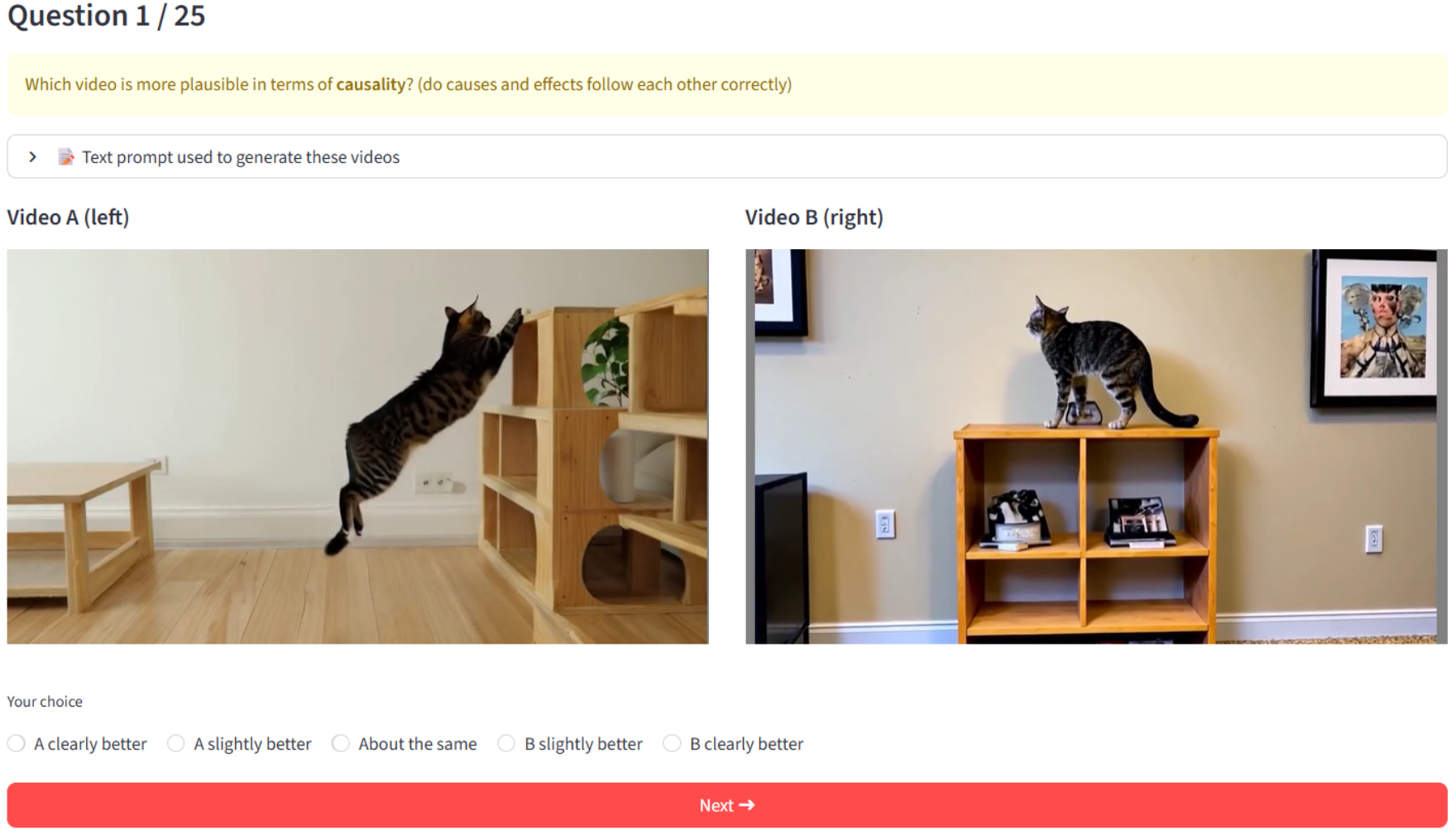}
    \captionof*{figure}{(b)}
\end{center}

\captionof{figure}{
\textbf{Website for human preference ranking.}
(a) Instructions and important notes for participants. (b) The comparing interface, where participants choose the video which perform better causal understanding.
}
\label{fig:user_study_web}

\noindent\textbf{Scoring and Aggregation.}
We first convert each judgment into a score $y \in \{-2, -1, 0, +1, +2\}$ that indicates how much video B is preferred over video A: \emph{A clearly better} $\rightarrow -2$, \emph{A slightly better} $\rightarrow -1$, \emph{About the same} $\rightarrow 0$, \emph{B slightly better} $\rightarrow +1$, and \emph{B clearly better} $\rightarrow +2$. We then use a \emph{cumulative-logit ordinal paired-comparison model}~\citep{tutz1986bradley,agresti1992analysis}, an extension of the Bradley--Terry model~\citep{bradley1952rank} that supports ties and graded preferences through ordered thresholds, to estimate a human preference score $s_m$ for each model $m$. We further extend it with participant-specific and prompt-specific scale terms and a position term, described below. These scores are not computed by a closed-form formula. Instead, they are the values that best explain the observed judgments, obtained in two steps: we first model the probability of each possible judgment, and then estimate the scores that make the observed judgments most likely.

\noindent\emph{Modeling a single judgment.}
Consider a comparison between model $a$ (shown as video A) and model $b$ (shown as video B), rated by participant $r$ on prompt $p$. The model first computes a preference strength
\begin{equation}
\eta = (s_b - s_a) \cdot \exp(w_r + \lambda_p) + \beta,
\end{equation}
which summarizes how strongly B is expected to be preferred over A in this particular comparison. The terms in $\eta$ are unknown parameters, defined as follows:
\begin{itemize}
    \item $s_b - s_a$ is the score gap between the two models, which is the quantity of interest.
    \item $w_r$ is a participant term that captures how strongly participant $r$ reacts to differences between videos. A participant with a small $w_r$ tends to answer $0$ even when the difference is noticeable, while a participant with a large $w_r$ tends to answer $\pm 2$. Modeling this prevents unusually cautious or decisive participants from distorting the model scores.
    \item $\lambda_p$ is a prompt term that captures how easily prompt $p$ reveals differences between models. For prompts whose causal interactions are easy to judge, $\lambda_p$ is larger, so the same score gap leads to stronger preferences.
    \item $\beta$ is a position term that captures any systematic preference for the left (A) or right (B) video, independent of content.
\end{itemize}
The preference strength $\eta$ is then converted into a probability for each of the five judgments using two thresholds $0 < c_1 < c_2$:
\begin{equation}
P(y \le k) = \sigma(\theta_k - \eta), \qquad (\theta_{-2}, \theta_{-1}, \theta_{0}, \theta_{+1}) = (-c_2, -c_1, c_1, c_2),
\end{equation}
where $\sigma(x) = 1/(1+e^{-x})$ is the logistic function and $P(y \le +2) = 1$. The probability of each individual judgment is the difference between consecutive cumulative probabilities; for example, $P(y = 0) = \sigma(c_1 - \eta) - \sigma(-c_1 - \eta)$. As a result, a large positive $\eta$ makes $+2$ most likely, a large negative $\eta$ makes $-2$ most likely, and $P(y = 0)$ is largest when $\eta = 0$, \ie when the two models have similar scores. The threshold $c_1$ determines where \emph{About the same} ends and \emph{slightly better} begins, and $c_2$ determines where \emph{slightly better} ends and \emph{clearly better} begins.

\noindent\emph{Estimating the parameters.}
All unknown parameters, namely the model scores $s_m$, the participant terms $w_r$, the prompt terms $\lambda_p$, the position term $\beta$, and the thresholds $c_1, c_2$, are estimated jointly from the 1,000 judgments (40 participants $\times$ 25 comparisons). We choose the values that maximize the total log-probability of the judgments actually observed:
\begin{equation}
\max_{\{s_m\}, \{w_r\}, \{\lambda_p\}, \beta, c_1, c_2} \; \sum_{t=1}^{1000} \log P(y_t \mid \eta_t) \; + \; \log P_{\text{prior}},
\end{equation}
where $y_t$ and $\eta_t$ are the judgment and preference strength of the $t$-th comparison. Intuitively, if B is frequently preferred over A, the optimization increases $s_b$ relative to $s_a$ so that the model assigns higher probability to these judgments. The prior term places weak Gaussian priors $s_m \sim \mathcal{N}(0, 1)$, $w_r, \lambda_p \sim \mathcal{N}(0, 0.35^2)$, and $\beta \sim \mathcal{N}(0, 0.25^2)$, and half-normal priors $c_1, c_2 - c_1 \sim \text{HalfNormal}(1)$. It keeps the estimates stable when data are limited (\eg when one model wins every comparison against another) and keeps $w_r$, $\lambda_p$, and $\beta$ close to zero unless the data indicate otherwise. Since only score differences affect $\eta$, the scores have no absolute zero point; we fix it by constraining the scores to have mean zero, which is why some models receive negative scores. We solve this optimization with a gradient-based method (L-BFGS). Once the scores are estimated, $s_b - s_a$ is simply the difference between the estimated scores of two models.

Judgments of $0$ are kept in the estimation: they indicate that the two models have similar scores, rather than being discarded or treated as half a win for each side. We rank the models by their estimated scores $s_m$ to obtain the overall human rank. This ranking closely agrees with the YoCausal ranking (CCI rank $+$ RSI rank, with ties broken by RSI rank), with a Kendall's $\tau$ of 0.667 ($p < 0.001$, exact permutation test). The complete ranking results are presented in~\cref{tab:human_preference_rank}.
\begin{table}[h]
    \centering
    \caption{
    \textbf{Overall ranking of human causality preference study.}
    All 13 models are ranked by their human preference score, estimated with the ordinal paired-comparison model from 1,000 pairwise judgments collected from 40 participants (higher is better). For reference, we also list each model's YoCausal rank (CCI rank $+$ RSI rank). The two rankings are strongly correlated (Kendall's $\tau = 0.667$, $p < 0.001$).
    }
    \label{tab:human_preference_rank}
    \resizebox{\textwidth}{!}{
    \begin{tabular}{l|ccccccc}
    \toprule
    Rank          & 1 & 2 & 3 & 4 & 5 & 6 & 7 \\
    \midrule
    Model         & Wan2.2-A14B & Wan2.1-14B & HunyuanVideo & Mochi & LTXV-2B & Wan2.2-5B & LTXV-13B \\
    Human Preference Score & 1.431 & 1.263 & 0.575 & 0.319 & 0.174 & 0.167 & 0.088 \\
    YoCausal Rank & 1 & 2 & 6 & 7 & 3 & 10 & 5 \\
    \midrule
    \midrule
    Rank          & 8 & 9 & 10 & 11 & 12 & 13 & \\
    \midrule
    Model         & CogVideoX-5B & Wan2.1-1.3B & CogVideoX1.5-5B & AnimateDiff-SD1.5 & CogVideoX-2B & AnimateDiff-SDXL & \\
     Human Preference Score & $-0.055$ & $-0.123$ & $-0.459$ & $-0.827$ & $-0.973$ & $-1.580$ & \\
    YoCausal Rank & 4 & 9 & 8 & 12 & 11 & 13 & \\
    \bottomrule
    \end{tabular}
    }
\end{table}

\subsection{Detailed Numerical Results}
\label{suppl_sec:numerical}

In the main paper, we primarily present the RSI and CCI results in graphical form to facilitate intuitive comparison of performance differences and trends across models. To enable precise numerical lookup and further analysis by the reader, this section provides the complete experimental data in tabular form. Specifically, ~\cref{tab:level1} reports the RSI scores of all 13 evaluated models and the human baseline across the four subsets ($\mathcal{D}_{\mathit{General}}$, $\mathcal{D}_{\mathit{Physics}}$, $\mathcal{D}_{\mathit{Human}}$, $\mathcal{D}_{\mathit{Animal}}$) as well as the overall dataset. ~\cref{tab:level2} provides the disaggregated RSI scores on the causal subset $\text{RSI}(\mathcal{D}_c)$ and the non-causal subset $\text{RSI}(\mathcal{D}_{nc})$, along with the resulting CCI scores. We additionally include the human baseline as a reference for both metrics.

\begin{table*}[h]
\centering
\caption{
\textbf{Numerical results of RSI.} This table shows RSI scores of all 13 evaluated models and the human baseline across four subsets ($\mathcal{D}_{\mathit{General}}$, $\mathcal{D}_{\mathit{Physics}}$
, $\mathcal{D}_{\mathit{Human}}$
, $\mathcal{D}_{\mathit{Animal}}$) and the overall average. Models are sorted by average RSI (RSI($\mathcal{D}$)) in ascending order. Bold values indicate the best-performing model in each column.
}
\begin{tikzpicture}
  \node[inner sep=0pt] (tbl) {%
    \resizebox{0.85\linewidth}{!}{%
      \begin{tabular}{@{}lcccccc@{}}
      \toprule
       & & \multicolumn{5}{c}{level 1: Reverse Surprise Index (RSI)$\uparrow$} \\
      \cmidrule(lr){3-7}
      & Release Date & $\mathcal{D}_{\mathit{General}}$ & $\mathcal{D}_{\mathit{Physics}}$ & $\mathcal{D}_{\mathit{Human}}$ & $\mathcal{D}_{\mathit{Animal}}$ & $\mathcal{D}$  \\
      \midrule
      AnimateDiff-SDXL & 04/2024
      & 27.80\% & 41.67\% & 48.73\% & 46.50\% & 41.18\% \\
      CogVideoX-2b & 08/2024
      & 33.20\% & 40.15\% & 56.64\% & 36.00\% & 41.50\% \\
      Wan2.1-T2V-1.3B & 03/2025
      & 29.80\% & 59.09\% & 59.15\% & 34.00\% & 45.51\% \\
      AnimateDiff-SD-1.5 & 06/2023
      & 33.00\% & 32.58\% & 61.68\% & 55.50\% & 45.69\% \\
      CogVideoX1.5-5b & 11/2024
      & 28.80\% & 62.12\% & 62.91\% & 33.50\% & 46.83\% \\
      Mochi-1-preview & 10/2024
      & 37.80\% & 43.18\% & 76.50\% & 39.00\% & 49.12\% \\
      CogVideoX-5b & 08/2024
      & 31.10\% & 67.42\% & 63.16\% & 38.00\% & 49.92\% \\
      Wan2.2-TI2V-5B & 07/2025
      & 34.40\% & 71.97\% & 63.75\% & 37.50\% & 51.91\% \\
      HunyuanVideo & 11/2025
      & 25.80\% & 64.39\% & \textbf{86.50\%} & 31.50\% & 52.05\% \\
      Wan2.1-T2V-14B & 03/2025
      & 37.60\% & 70.45\% & 66.92\% & 38.00\% & 53.24\% \\
      Wan2.2-T2V-A14B & 07/2025
      & 36.80\% & \textbf{77.27\%} & 66.17\% & 36.50\% & 54.19\% \\
      LTX-Video-13b-0.9.8 & 07/2025
      & \textbf{61.20\%} & 47.73\% & 47.50\% & \textbf{69.50\%} & 56.48\% \\
      LTX-Video-2b-0.9.6 & 04/2025
      & 58.60\% & 57.58\% & 54.25\% & 65.00\% & \textbf{58.86\%} \\
      \midrule
      Human & 
      & 76.60\% & 91.7\% & 76.0\% & 72.0\% & 79.08\% \\
      \bottomrule
      \end{tabular}%
    }%
  };
  \draw[->, thick, gray, overlay]
    ([xshift=-6pt, yshift=-4.5em] tbl.north west) --
    ([xshift=-6pt, yshift= 2.5em] tbl.south west);
  \node[anchor=south, font=\scriptsize\itshape, gray, align=center, overlay]
    at ([xshift=-25pt, yshift= 0em] tbl.south west) 
    {
    Better \\ 
    perceives \\ 
    arrow of time
    };
    
\end{tikzpicture}
\label{tab:level1}
\end{table*}

\begin{table*}[h]
\centering
\caption{
\textbf{Numerical results of CCI.}
This table shows CCI scores of all 13 evaluated models and the human baseline, along with the disaggregated RSI on the causal subset $\text{RSI}(\mathcal{D}_c)$ and non-causal subset $\text{RSI}(\mathcal{D}_{nc})$. Normalized CCI rescales raw CCI values relative to the human baseline. Models are sorted by CCI in ascending order. Bold values indicate the best-performing model in each column.
}
\begin{tikzpicture}
  \node[inner sep=0pt] (tbl) {
    \resizebox{0.85\linewidth}{!}{
      \begin{tabular}{@{}lccccc@{}}
      \toprule
        & & \multicolumn{4}{c}{level 2: Causality Cognition Index (CCI) $\uparrow$} \\
      \cmidrule(lr){3-6}
      & Release Date & CCI($D$) $\uparrow$ &  Normalized CCI($D$) & RSI($D_c$) & RSI($D_{nc}$) \\
      \midrule
        
        AnimateDiff-SD-1.5 & 06/2023
        & -5.21\% & -60.09\% & 43.40\% & 48.61\% \\
        AnimateDiff-SDXL & 04/2024
        & -5.07\% & -58.48\% & 38.93\% & 44.00\% \\
        LTX-Video-13b-0.9.8 & 07/2025
        & -4.32\% & -49.83\% & 54.65\% & \textbf{58.97\%} \\
        Wan2.2-TI2V-5B & 07/2025
        & -2.12\% & -24.45\% & 50.90\% & 53.02\% \\
        HunyuanVideo & 11/2025
        & -0.29\% & -3.34\% & 51.15\% & 51.44\% \\
        LTX-Video-2b-0.9.6 & 04/2025
        & -0.20\% & -2.31\% & \textbf{57.95\%} & 58.15\% \\
        CogVideoX-2b & 08/2024
        & 0.93\% & 10.73\% & 41.11\% & 40.18\% \\
        Mochi-1-preview & 10/2024
        & 3.85\% & 44.41\% & 49.11\% & 45.26\% \\
        CogVideoX1.5-5b & 11/2024
        & 4.85\% & 55.94\% & 48.46\% & 43.61\% \\
        CogVideoX-5b & 08/2024
        & 5.09\% & 58.71\% & 51.36\% & 46.27\% \\
        Wan2.1-T2V-1.3B & 03/2025
        & 5.36\% & 61.82\% & 46.92\% & 41.56\% \\
        Wan2.2-T2V-A14B & 07/2025
        & 5.51\% & 63.55\% & 55.73\% & 50.22\% \\
        Wan2.1-T2V-14B & 03/2025
        & \textbf{5.91\%} & \textbf{68.17\%} & 54.80\% & 48.89\% \\
      \midrule
      Human & 
      & 8.67\% & 100.00\% & 85.09\% & 76.42\% \\
      \bottomrule
      \end{tabular}%
    }
  };
  \draw[->, thick, gray, overlay]
    ([xshift=-6pt, yshift=-4.5em] tbl.north west) --
    ([xshift=-6pt, yshift= 2.5em] tbl.south west);
  \node[anchor=south, font=\scriptsize\itshape, gray, align=center, overlay]
    at ([xshift=-25pt, yshift= 0em] tbl.south west) 
    {
    Better \\ 
    understand \\ 
    causality
    };
    
\end{tikzpicture}
\label{tab:level2}
\end{table*}

\section{Scaling Law and Generational Evolution in Causal Cognition}
\label{suppl_sec:scaling_law}

\begin{figure*}[h]
    \centering
    \includegraphics[width=0.85\linewidth]{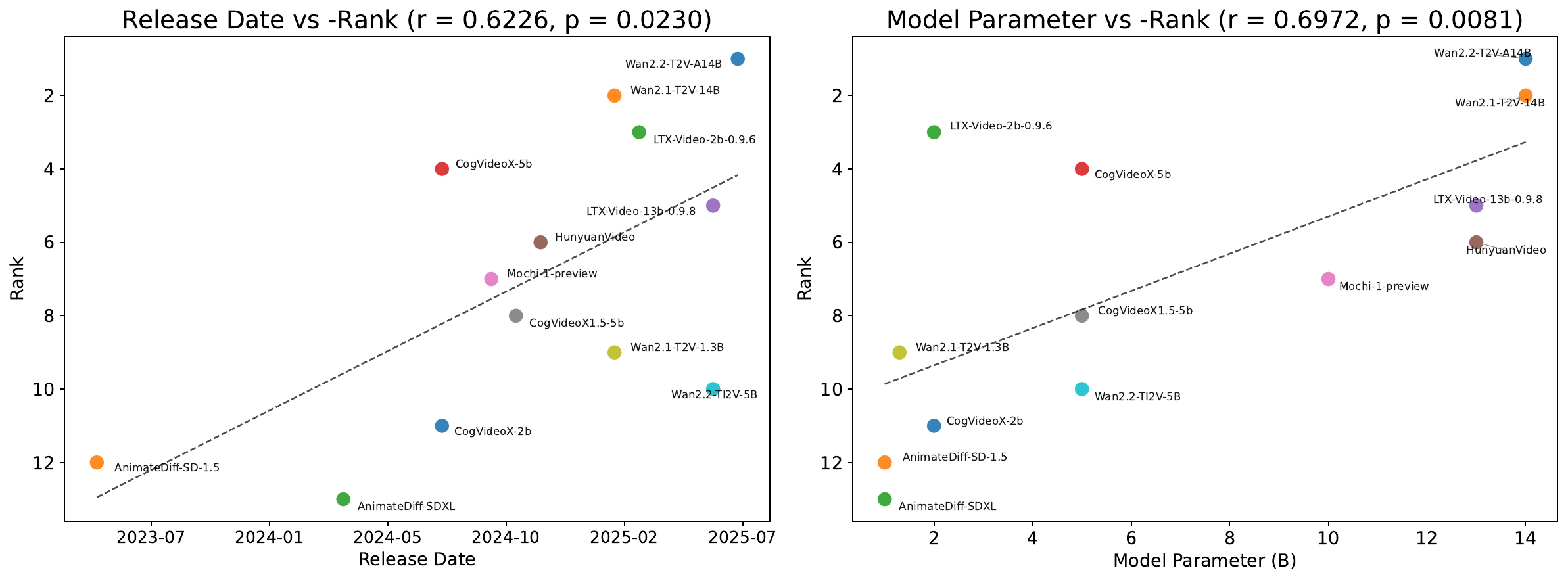}
    \caption{
    \textbf{Scaling laws and generational trends in causal cognition.} Aggregate causal cognition rank correlates positively with both release date (Pearson's $r{=}0.6226$) and parameter count (Pearson's $r{=}0.6972$), indicating that larger and newer models exhibit stronger causal understanding. 
    }
    \label{fig:para_date_rank}
\end{figure*}

Within the prevailing research trajectory of advancing video generation models toward world models, scaling laws have been regarded as a guiding principle~\citep{yin2025towards,liang2024scaling,kaplan2020scaling}. We therefore investigate whether model size and release date correlate with causal cognition by computing the correlation between these factors and our aggregate rank. As shown in \cref{fig:para_date_rank}, both release date and model parameters exhibit significant correlations with the aggregate ranking (Pearson's $r = 0.6226$ and Pearson's $r = 0.6972$, respectively). This demonstrates that increasing parameter count does enhance causal understanding to a meaningful degree, validating the effectiveness of scaling laws for this higher-level cognitive capability. Furthermore, models have progressively improved in causal cognition across generations. For instance, the transition from UNet-based architectures to DiT-based architectures has yielded substantial improvements in causal perception for most model families.

\clearpage

\bibliographystyle{abbrv}
\bibliography{main}

\end{document}